\documentclass[journal]{IEEEtran}
\usepackage{balance}
\usepackage{amsfonts}
\usepackage{float}
\usepackage[pdftex]{graphicx}
\usepackage{subfigure}
\usepackage{epstopdf}
\usepackage[colorlinks,linkcolor=red]{hyperref}
\ifCLASSOPTIONcompsoc
\usepackage{booktabs}
  \usepackage[nocompress]{cite}
\else
  \usepackage{cite}
\fi
\usepackage{algorithm}
\usepackage{algorithmic}
\usepackage{amsmath}
\usepackage{makecell}
\usepackage{multirow}
\usepackage{color}
\usepackage{graphicx}
\usepackage{float}
\usepackage{booktabs}
\usepackage{bigstrut}

\begin{document}

\title{Semantic Interleaving Global Channel Attention for Multilabel Remote Sensing Image Classification}
\newcommand{\w}{\textcolor{black}}
\newcommand{\s}{\textcolor{red}}
\author{Yongkun Liu,~
        Kesong Ni,~
        Yuhan Zhang,~
        Lijian Zhou,~
        Kun Zhao*       

\thanks{Corresponding author: Kun Zhao. E-mail: sterling1982@163.com}
\thanks{Yongkun Liu, Kesong Ni, Yuhan Zhang, Lijian Zhou and Kun Zhao were with the School of Information and Control Engineering, Qingdao University of Technology, Qingdao 266520, China. }
}
\markboth{YONGKUN LIU \MakeLowercase{\textit{et al.}}: Semantic Interleaving Global Channel Attention for Multilabel Remote Sensing Image Classification}
{Yongkun Liu \MakeLowercase{\textit{et al.}}:Semantic Interleaving Global Channel Attention for Multilabel Remote Sensing Image Classification}
\maketitle

\begin{abstract}
Multi-Label Remote Sensing Image Classification (MLRSIC) has received increasing research interest. Taking the cooccurrence relationship of multiple labels as additional information helps to improve the performance of this task. Current methods focus on using it to constrain the final feature output of a Convolutional Neural Network (CNN). On the one hand, these methods do not make full use of label correlation to form feature representation. On the other hand, they increase the label noise sensitivity of the system, resulting in poor robustness. In this paper, a novel method called “Semantic Interleaving Global Channel Attention” (SIGNA) is proposed for MLRSIC. First, the label co-occurrence graph is obtained according to the statistical information of the data set. The label co-occurrence graph is used as the input of the Graph Neural Network (GNN) to generate optimal feature representations. Then, the semantic features and visual features are interleaved, to guide the feature expression of the image from the original feature space to the semantic feature space with embedded label relations. SIGNA triggers global attention of feature maps channels in a new semantic feature space to extract more important visual features. Multihead SIGNA based feature adaptive weighting networks are proposed to act on any layer of CNN in a plug-and-play manner. For remote sensing images, better classification performance can be achieved by inserting CNN into the shallow layer. We conduct extensive experimental comparisons on three data sets: UCM data set, AID data set, and DFC15 data set. Experimental results demonstrate that the proposed SIGNA achieves superior classification performance compared to state-of-the-art (SOTA) methods. It is worth mentioning that the codes of this paper will be open to the community for reproducibility research.
Our codes are available at {\small\url{https://github.com/kyle-one/SIGNA}}.
\end{abstract}


\IEEEpeerreviewmaketitle

\section{Introduction}

\begin{figure}[!htb]
	\centering
	\includegraphics[width=1.0\linewidth]{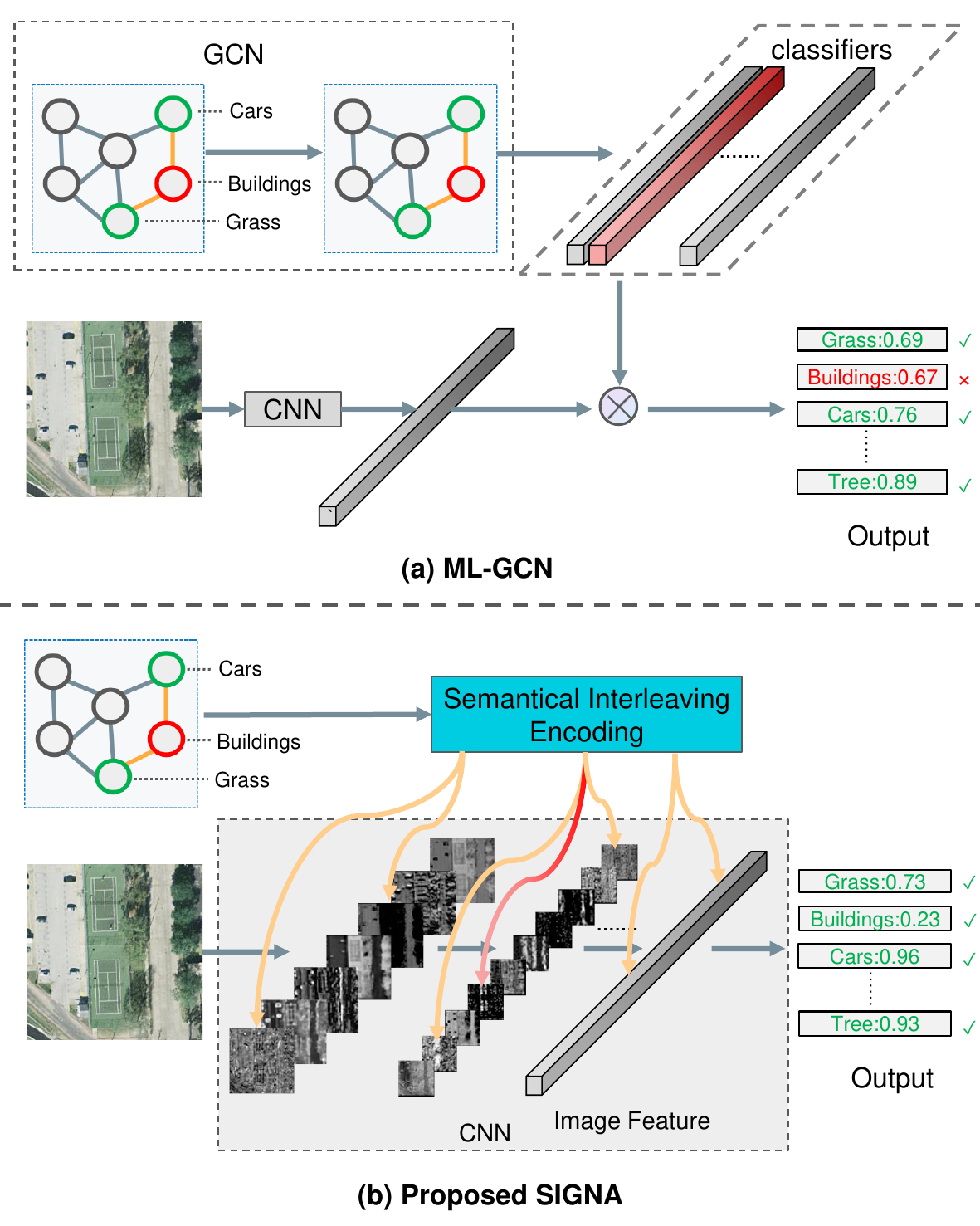}
	\caption{
	Subfigure (a) explains the basic idea of ML-GCN [1]. This method builds object classifiers using GCN. These object classifiers are used explicitly for the final output features of the CNN. 
Subfigure (b) explains the basic idea of the proposed SIGNA. Using the label co-occurrence graph, the visual shallow features are semantical interleaving encoded. 
As shown in the input label co-occurrence graph, when cars and grass exist, there is a high probability that buildings exist. 
In subfigure (a), this prior knowledge does not match the current example, leading to wrong inferences.
While in subgraph (b), the mislabeled relations are diluted by the semantic interleaving encoding, which improves the robustness.
	}
	\label{fig:Motivations}
\end{figure}

\IEEEPARstart{W}{ith} the development of satellite technology, remote sensing image scene classification has become a very active field.
So far, remote sensing images are widely used in various problems, such as urban cartography\cite{zhu2019so2sat,hong2019learning,rasti2020feature,hong2019learnable,hua2020relation,gao2017optimized}, land use determination\cite{tan2017multi,hua2019recurrently,wang2019multi}, terrain surface analysis\cite{tan2017multi} and spectral unmixing\cite{fernandez2020endmember,hong2018augmented,fernandez2018hyperspectral}.
Remote sensing image classification plays a key role in the above tasks and has become one of the basic tasks in the remote sensing field.
Single label remote sensing image classification aims to predict a semantic classification based on the overall information in the remote sensing image\cite{lin2013cross,movshovitz2015ontological}. 
However, because the information contained in remote sensing images is too rich and complicated, it is difficult to describe an image with a single label at the macro level.  
Unlike the single label used in traditional remote sensing classification, MLRSIC uses a series of object labels to describe remote sensing images.
This is more meaningful for semantic description and visual understanding of remote sensing images.

\begin{figure}[t]
	\centering
	\includegraphics[width=1.0\linewidth]{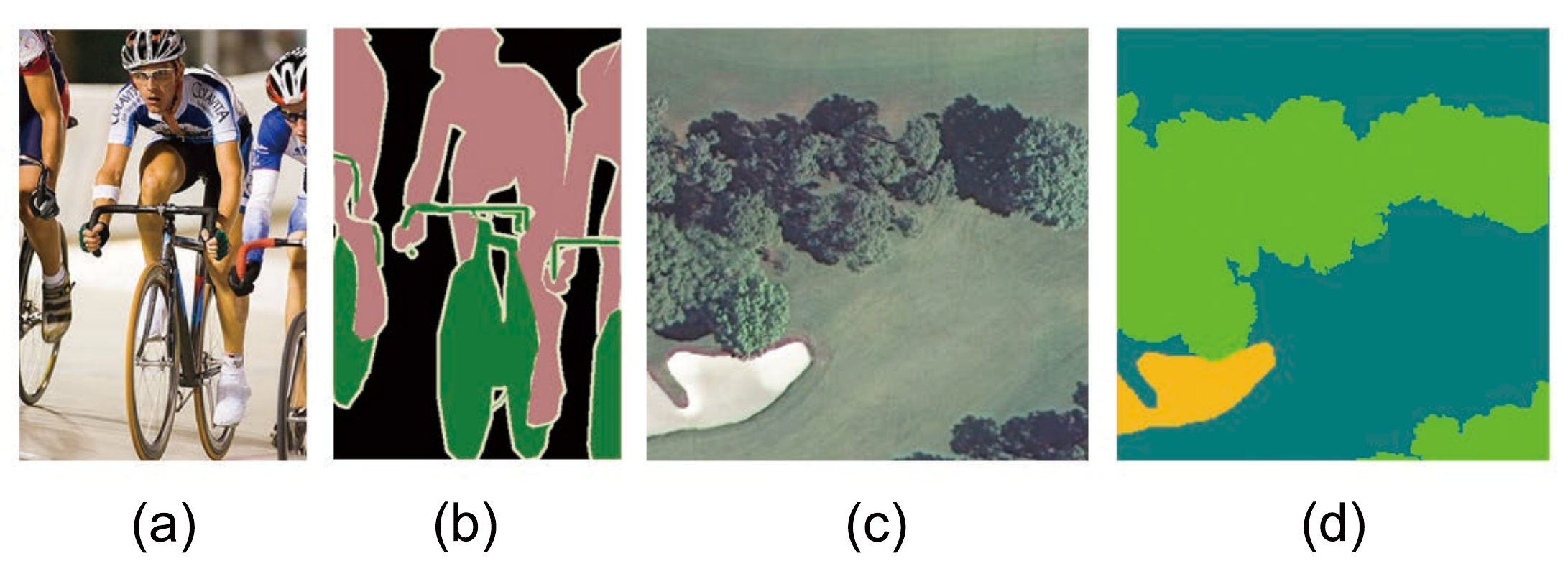}
	\caption{(a), (b) are the sample images in general classification tasks and their corresponding semantic segmentation maps. 
(c), (d) are sample images in remote sensing and their corresponding semantic segmentation maps.
	}
	\label{fig:ex}
\end{figure}

\subsection{Motivation}
\label{subsec:Motivation}

Studying how to apply label correlation to multi-label image classification is critical for improving MLRSIC performance. Recently, some related methods have been proposed.
Hua et al.\cite{hua2019recurrently} proposed a method to extract image features through CNN backbone network, which connects and sends the image features to bidirectional LSTM for classification, and implicitly learn label co-occurrence information.     
Chen et al.\cite{chen2019multi} proposed the 
ML-GCN method. By modeling the label correlation as a graph, the GNN is used to process the input image. 
Koda et al.\cite{koda2018spatial} proposed a multi-label classification method based on structured support vector machine (SVM) using the relationship between labels.
All of the methods above employ CNN to extract visual features, then use various methods to fuse label features and visual features in order to apply label correlation to image classification.
This feature fusion method has some drawbacks.
As shown in (a) of Fig.\ref{fig:Motivations}, the vector of the output of the CNN and several classifier vectors are mapped to one output.
Above all, semantic features are explicitly used to guide image feature classification, which results in the networks being highly sensitive to the numerical value and noise of label features.
Furthermore, from the perspective of back propagation, the image features and semantic features exchange gradients only once in the last layer, which indicates that the semantic features are not fully utilized.
Last but not least, these methods do not take into account the properties of remote sensing images.
As shown in Fig. \ref{fig:ex}, there are two significant distinctions between images in remote sensing classification tasks and images in general visual classification tasks.
First, generic image semantic labels correspond to more abstract visual representations. In (a), the image pixels corresponding to people have yellow skin, blue-white-black clothes, black pants, and white-black helmets, which are very complex.
However, the pixels corresponding to the semantic labels of remote sensing images have stronger regional consistency in texture and color, so it can be speculated that they rely more on shallow feature representations.  In (b), the image pixels corresponding to the tree labels have uniform green color and similar texture.
Second, the semantic labels of generic images only correspond to part of the pixels, because there is a lot of useless background information.

\subsection{Contributions}
Based on the above three problems, the correlation between semantic label co-occurrence relationships and visual features in remote sensing images has been intensively studied.
\begin{itemize}
\item Semantic interleaving encoding is proposed, inspired by interleaved codes in error correction codes.  This is to solve the problem of poor network robustness caused by explicit use of semantic features. It can reduce the numerical and noise sensitivity of fixed label co-occurrence relationships, thereby improving the robustness of the entire network.

\item SIGNA is proposed, for the problem that semantic features are not fully utilized. Global attention to image channels is triggered in a new semantic feature space, and the more significant visual features are extracted by using relationships between labels and channels.

\item A plug-and-play network based on SGINA is designed to make better use of the characteristic of remote sensing image data sets.
The network is used in the shallow layer of CNN. On the one hand, it can make better use of the texture and edge features of remote sensing images. On the other hand, it does not ignore any meaningful pixel in the shallow layer.

Extensive experiments on AID, UCM, and DFC15 data sets show that our method is optimal compared to several other methods that use label constraints. The relevant code of this article will be publicly released for reproducible research in the community.
\end{itemize}

\subsection{Section Arrangement}

The rest of the paper is organized as follows. In Section~\ref{sec:Related Work}, we review the multi-label image classification algorithms in remote sensing, the current research progress of attention mechanism in multi-label classification, and introduce the application of GNN in the direction of image recognition.
Section~\ref{sec:Proposed Approach}  introduces the proposed SIGNA and describes in detail its three modules. Section~\ref{sec:Experiments and Analysis} reports the experimental setup, results and discussions. Section~\ref{sec:Conclusion and Future Work} concludes the paper.

\section{Related Work}
\label{sec:Related Work}

The MLRSIC task is a hot research topic in the field of remote sensing recently, which can provide a more fine-grained understanding of images. 
We review related work from the following aspects: traditional methods for MLRSIC, using label relationships and attention mechanisms, GNN applications in MLRSIC.

\subsection{Multi-label Classification In Remote Sensing}
\label{subsec:Multi-label classification in remote sensing}
In the early research of multi-label classification, hand-crafted features are often used (such as color, texture, visual word bag), describe the image scene and combine it with traditional machine learning methods, such as random forest and support vector machine (SVM). Geng et al.\cite{geng2015high} proposed a deep convolutional AE (DCAE) to extract features and conduct classification automatically. 
Dai\cite{dai2018novel} used raw pixel values, simple bag of spectral values and the extended bag of spectral values descriptors to characterize the spectral content.
Koda et al.\cite{koda2018spatial} used inter-label relations by means of a structured SVM and incorporated spatial contextual information by improving cost function.
Zeggada et al.\cite{zeggada2018multilabel} used a multilabel CRF model to  integrate spatial correlation between adjacent tiles and the correlation between labels within the same tile, in order to improve the multilabel classification map associated with the input image. 
However, the generalization ability of above research is limited, and it is difficult to express the overall and high-level semantic information. Therefore, the classification performance of these algorithms is limited.

In the field of computer vision, with the establishment of large-scale natural image data sets and the rapid development of deep convolution networks, such as VGG16\cite{simonyan2014very}, ResNet-50\cite{he2016deep}, GoogleNet\cite{szegedy2015going} and other networks, the research of multi-label image classification has attracted attention and developed rapidly. Deep learning methods such as CNN are also widely used in remote sensing image scene classification. 
Zhao\cite{zhao2016spectral} proposed a spectral-spatial-feature-based classification framework, which jointly makes use of a local-discriminant embedding-based dimension-reduction algorithm and a 2-D CNN. Ma\cite{ma2021scenenet} proposed a framework for scene classification network architecture search based on multi-objective neural evolution.

Although CNN has achieved remarkable results in multi-label remote sensing image classification, there is still room for improvement. First, for remote sensing images, objects of different labels are of different sizes and dispersed, making it difficult for CNN to identify important label objects. An attention mechanism can be used to address this problem by ignoring irrelevant information in order to focus on important information. Second, the number of predicted standard combinations grows exponentially with the number of categories. To solve this problem, the introduction of tag correlation can effectively control the size of the tag space.

\subsection{Using Multi-label Relationships and Attention Mechanisms}
\label{subsec:Using Multi-label Relationships And Attention Mechanisms}

Introducing label correlation and attention mechanisms into CNN has been shown to be an effective way to improve multi-label image classification.
In recent years, some methods have been proposed to apply attention mechanism in remote sensing image classification.

Tong et al.\cite{tong2020channel} proposed a channel-attention-based DenseNet (CAD) network for scene classification.  
Li et al.\cite{li2020augmentation} proposed an attention mechanism-based CNN with multi-augmented schemes.
Yu et al.\cite{yu2020hierarchical} proposed a feature fusion framework based on hierarchical attention and bilinear pooling  for the scene classification of remote sensing images.

In the MLRSIC task, there are the following studies that use label correlation to improve network performance.
In the CNN-RNN framework proposed by Wang\cite{wang2016cnn}, the RNN model learns joint image-label embeddings from CNN features, and uses the memory mechanism of RNN to predict labels in an ordered prediction path. 
Hua et al. \cite{hua2019recurrently} proposed a method to extract image features through the CNN backbone network, then connect and send the image features to a bidirectional LSTM for classification, and implicitly learn label co-occurrence information. 
Zhang et al.\cite{zhang2017dual} put the label co-occurrence matrix into two convolutional layers and two fully connected layers to learn label correlations.
Huang et al.\cite{huang2021multilabel} proposed an end-to-end deep learning framework consisting of a multi-scale feature fusion module, a channel-spatial attention learning module, and a label correlation extraction module to effectively exploit the correlations between multiple labels. 
Bazi et al.\cite{bazi2019two} proposed a dual-branch neural network composed of an image branch and a label branch for remote sensing image classification, and improved the loss function related to image label similarity and label discrimination.
Zhu et al.\cite{zhu2020deep} adopted a two-layer semantic concept to annotate multi-label RS images, and used a classification branch for multi-label annotation and an embedding branch to preserve scene-level similarity relations for classification.

The label relationship constraints in the above studies mostly use various methods to constrain the output features of CNN, and do not affect the internal parameters of CNN. The attention mechanism embedded inside CNN does not introduce label correlation.

\subsection{GNN In Remote Sensing}
\label{subsec:Graph neural network in remote sensing}

\begin{figure*}[ht]
	\centering
	\includegraphics[width=1.0\linewidth]{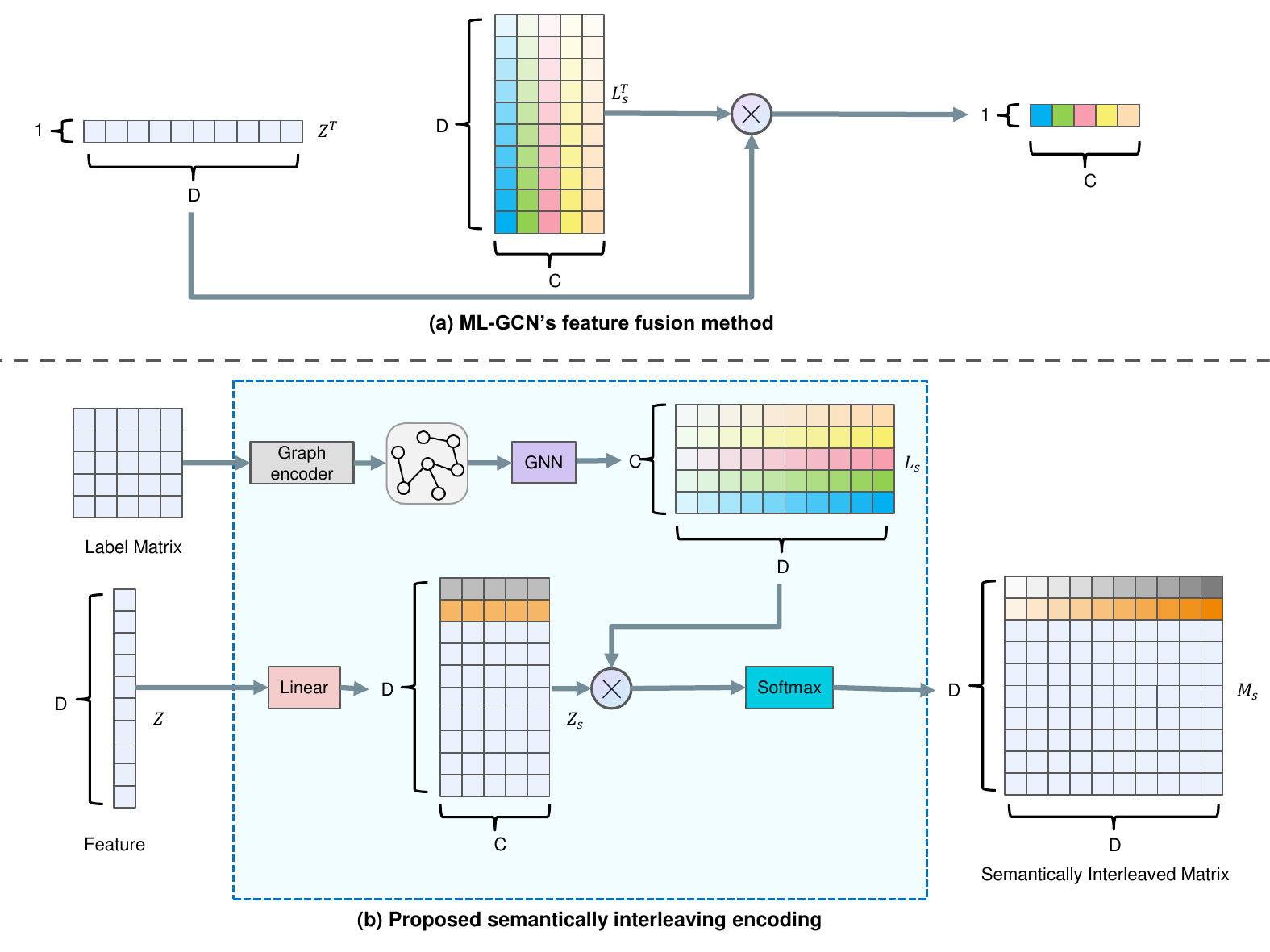}
	\caption{Pipeline for semantic interleaving encoding. (a) The calculation process of ML-GCN. (b) The calculation process of semantic interleaving encoding.}
	\label{fig:encoding}
\end{figure*}

\begin{figure*}[ht]
	\centering
	\includegraphics[width=1.0\linewidth]{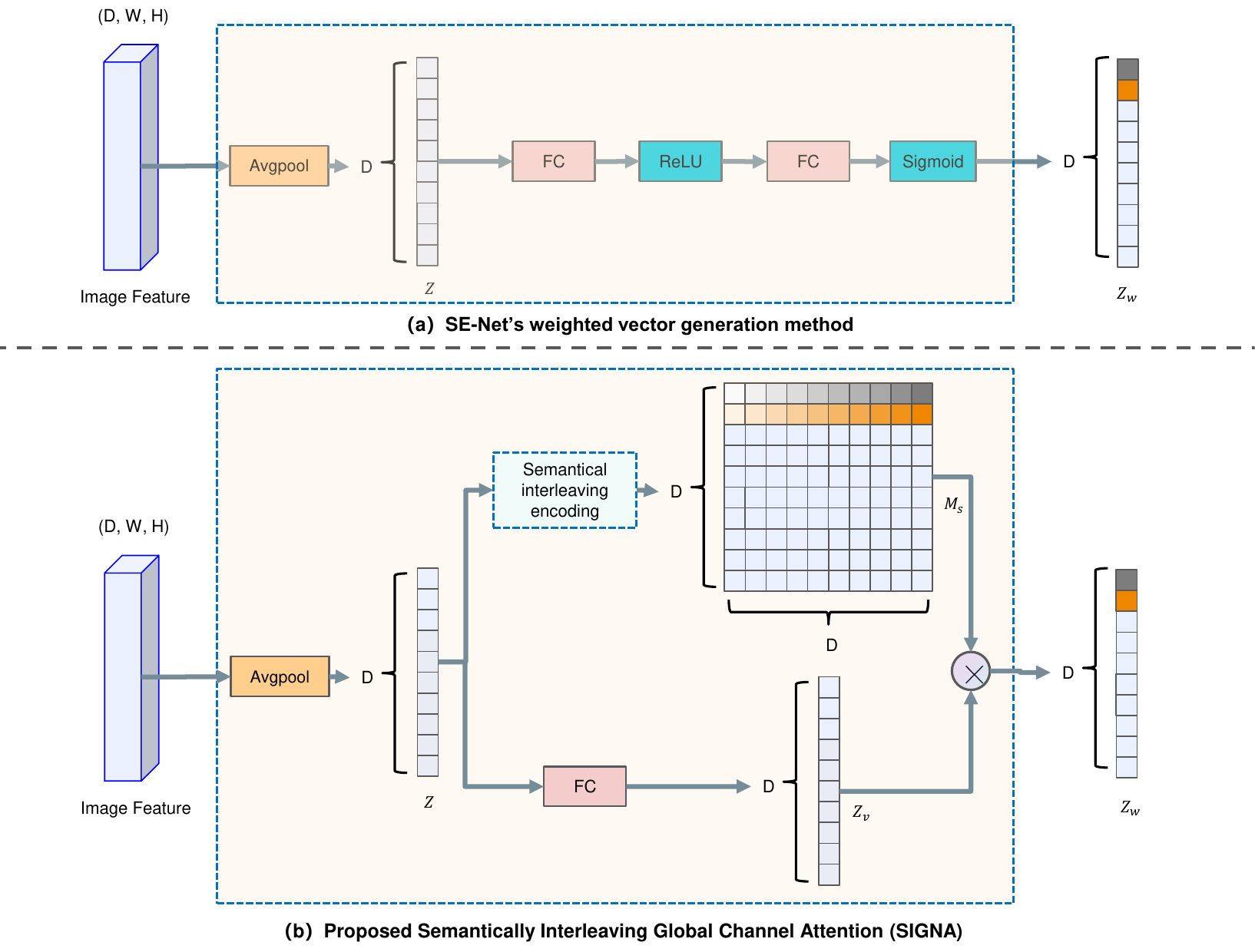}
	\caption{Comparison of SIGNA and SE-Net calculation methods.}
	\label{fig:Attention}
\end{figure*}

GNN is a new hot research direction in the field of artificial intelligence. It is a network for graph structure data, which can implicitly learn the spatial and topological relationships of graphs. Recently, many high-performance GNN models have been developed, such as Graph Convolution Network (GCN)\cite{kipf2016semi},  Graph Attention Network (GAT)\cite{velickovic2017graph}, Graph SAmple and agreGatE (GraphSAGE)\cite{hamilton2017inductive}, etc. When GNN is used in image recognition, some strategy is usually used to construct nodes and edges as the input of GNN. The node information in GNN can be transmitted in the graph, and the characteristics of nodes in the graph can be aggregated.

In the field of remote sensing, there are several methods to construct scene map. Kang et al.\cite{kang2020graph} proposed graph relation network (GRN) method, which uses GNN to model multi-label semantic proximity and learn the graph-based neighborhood semantic relationship between multi-label remote sensing images in metric space. Li et al.\cite{li2020multi} proposed a method to build a scene graph for each image, generate high-level appearance features from the image as the initial node of the graph, and mine the topological relationship of the graph by using the adaptive learning ability of GAT. Liang et al.\cite{liang2020deep}
regard the ground object as a node, use the detector to detect the object, and then define the adjacent relationship between nodes according to the spatial distance between entities.
Lin et al. \cite{lin2021multilabel} constructed a concept map for the whole label set by using human knowledge in conceptnet\cite{speer2017conceptnet}, and then fused the two global feature vectors generated by CNN and GNN. In this method, the feature elements of GCN output and CNN output are multiplied, and then 
output to the final classifier for scene classification. In this strategy, the original CNN output is replaced by the output generation of feature fusion. 
However, this explicit feature fusion is highly sensitive to noise in label features. This results in poor fusion and system robustness.

\section{Proposed Approach}
\label{sec:Proposed Approach}

In this part, we elaborate on our proposed Semantic Interleaving Global Channel Attention (SIGNA) for MLRSIC in detail.

\subsection{Semantical Interleaving Encoding}
\label{subsub:semantic interleaving Encoding}

Label relationship is obtained from overall statistics of the data set. So for each image instance, this label relationship is noisy.
Semantic interleaving coding is proposed to reduce noise sensitivity when exploiting label relations, which is inspired by interleaved codes in error correction codes.
Interleaved codes are used to transform a burst channel into a random independent error channel by encoding n pieces of original data of length m into m pieces of data of length n (i.e: write in rows and read out in columns). It is used to tackle the channel's continuous error problem and increase the coding's robustness and reliability.
Semantic interleaving encoding follows the similar concept.
It is performed in two steps. The details of its implementation are detailed below.

\subsubsection{Using GNN to model label co-occurrence matrix as semantic features}
\label{subsub:Using GNN to model label co-occurrence matrix as discrete label relation features}

By counting the probability of occurrence of label pairs, a co-occurrence matrix can be obtained. Specifically, it is to traverse the labels of each picture, and count the number $N_{ij}$ of $L_j$ occurrences when the label $L_i$ appears. Calculate the probability matrix by 
\begin{equation}\label{eq:coo}
    P_{ij}=N_{ij}/N_{ii},(i,j<C),
\end{equation}
where $N_{ii}$ represents the total number of occurrences of label $i$, $C$ represents the number of types of labels, and all $P_{ij}$ constitute a co-occurrence matrix.
In order to avoid the smoothing phenomenon of the GNN, Formula \ref{eq:smoothing} is used to process the co-occurrence matrix.

\begin{equation}\label{eq:smoothing}
    \boldsymbol{G}_{i j}=\left\{\begin{array}{ll}
0, & \text { if } \boldsymbol{P}_{i j}<Q \\
{P}_{i j}, & \text { if } \boldsymbol{P}_{i j} \geq Q
\end{array}\right.
\end{equation}
where the threshold $Q$ is set to 0.4.

After the initial label co-occurrence matrix is constructed, the popular GCN among GNNs is chosen to model the label co-occurrence matrix.
Graph $G$ is used as the relation node of GCN. 
The Glove word embedding vector of the label word vector is used as input. 
After GCN, the original label relationship graph $G \in \mathbb{R}^{C \times C}$ is encoded as semantic feature $L_s \in \mathbb{R}^{C \times D}$ ($D$ usually selects the number of channels of the feature map, $C$ is the number of categories of the data set labels), thereby dispersing the label relationship into $D$ channels.

GCN updates nodes through information propagation between nodes. The goal of a GCN is to learn a function $f( \cdot, \cdot )$ on the spatial domain of a graph $G$, whose input is the feature description $H^l \in \mathbb{R}^{n \times d} $ and the corresponding correlation matrix $G \in \mathbb{R}^{n \times n}$. $l$ is the number of layers of the GCN, $n$ is the number of nodes, and $d$ is the dimension of each node feature.
In the task of modeling semantic features, $n$=$C$, and $d$=$D$ for the last layer of GCN.
That is, through GCN, the final discrete label relationship feature $L_s \in \mathbb{R}^{C \times D}$ is output of GCN $H^{l+1}$.
Therefore, the node feature output learned by 
the GCN is denoted as $H^{l+1} \in \mathbb{R}^{n \times d'}$. Each GCN layer can be represented by a nonlinear function f: 

\begin{equation}\label{eq:GCN1}
H^{l+1} = f (H^{l} , G)
\end{equation}

Next, the convolution operation in \cite{kipf2016semi} is introduced, and the $f(\cdot ,\cdot )$ function can be further expressed as:

\begin{equation}\label{eq:GCN2_4}
\begin{aligned}
H^{l+1} &=\delta\left(\hat{A} H^{l} W^{l}\right) \\
\hat{G} &=\tilde{D}^{-1 / 2} \tilde{G} \tilde{D}^{-1 / 2} \\
\tilde{D_{i i}} &=\sum_{j} \tilde{G}_{i j} \\
\tilde{G} &=G+I_{K}
\end{aligned}
\end{equation}

where $W^l \in \mathbb{R}^{d \times d}$ denotes a learnable transformation matrix, $\hat{G}$ indicates the normalized representation of $G$, $I_{K}$ is an identity matrix, $\delta (\cdot)$ is an activation function, and we set it as LeakyReLU function. 
In this way, we can learn and model the interrelationships between nodes by stacking multiple layers of GCNs. 
Similar to the above methods, the GCN part of the Label Graph Relation Encoder can be replaced by the recently proposed SAGE \cite{hamilton2017inductive} and GAT\cite{velivckovic2017graph}. 
The detailed algorithm can be found in their paper.  In the experimental section, we detail the differences between the three GNN for modeling.

\subsubsection{Interleaving semantic feature and image feature}
\label{subsub:Interleaving semantic features and image features}
At this point, the semantic feature $L_s \in \mathbb{R}^{C \times D}$ has been obtained, and then it will be interleaved with the visual feature. Thereby, the image global features are guided into the semantic feature space with label correlation. 

As shown in Fig.\ref{fig:encoding}(b), given a global feature vector $Z$ of length $D$, in order to facilitate interleaving, a linear transformation and reshape is first passed to obtain the expanded feature $Z_s \in \mathbb{R}^{D \times C}$. 

The semantic feature matrix $L_s$ is matrix-multiplied with the extended image global feature matrix $Z_s$.
The real domain of the result is then mapped to the [0,1] space through the softmax function. Finally, the semantic interleaving matrix $M_s$ is obtained.
The calculation formula is:
\begin{equation}\label{eq:encoding}
M_s=softmax(reshape(ZA^T+b) L_s)
\end{equation}
where reshape adjusts the number of rows and columns of $Z$ after the linear transformation, resulting in a matrix $Z_s \in \mathbb{R}^{D \times C}$. $A^T$ denotes a learnable transformation matrix, and $b$ is a learnable bias.    

Each row of the $L_s$ matrix represents a feature vector, and each column of the $Z_s$ matrix represents a discrete label correlation feature vector. Each row of the $M_s$ matrix represents the distribution of an eigenvalue of the image global feature vector $Z$ in the discrete semantic feature space.

It is worth noting that our multiplication method for semantic interleaving is different from that of methods such as ML-GCN.
In ML-GCN, matrix multiplication of $Z \in \mathbb{R}^{1 \times W}$ and $L_s^T \in \mathbb{R}^{W \times C}$  is performed to obtain the final classification $Out \in \mathbb{R}^{1 \times C}$.(Fig.\ref{fig:encoding}(a))
This is equivalent to using the output matrix $L_s$ of GCN as $C$ classifiers, explicitly using semantic features for classification.
In contrast, we perform matrix multiplication of $Z_s \in \mathbb{R}^{W \times C}$ and $L_s \in \mathbb{R}^{C \times W}$ to get the semantic interleaving matrix $M_s \in \mathbb{R}^{W \times W}$. This implicitly interleave semantic and visual features, thereby implicitly guiding image global features into a semantic feature space that is highly correlated with labels.(Fig.\ref{fig:encoding}(b))
And this reduces the noise sensitivity of semantic features as explicit classifiers.

\begin{figure*}[ht]
	\centering
	\includegraphics[width=1.0\linewidth]{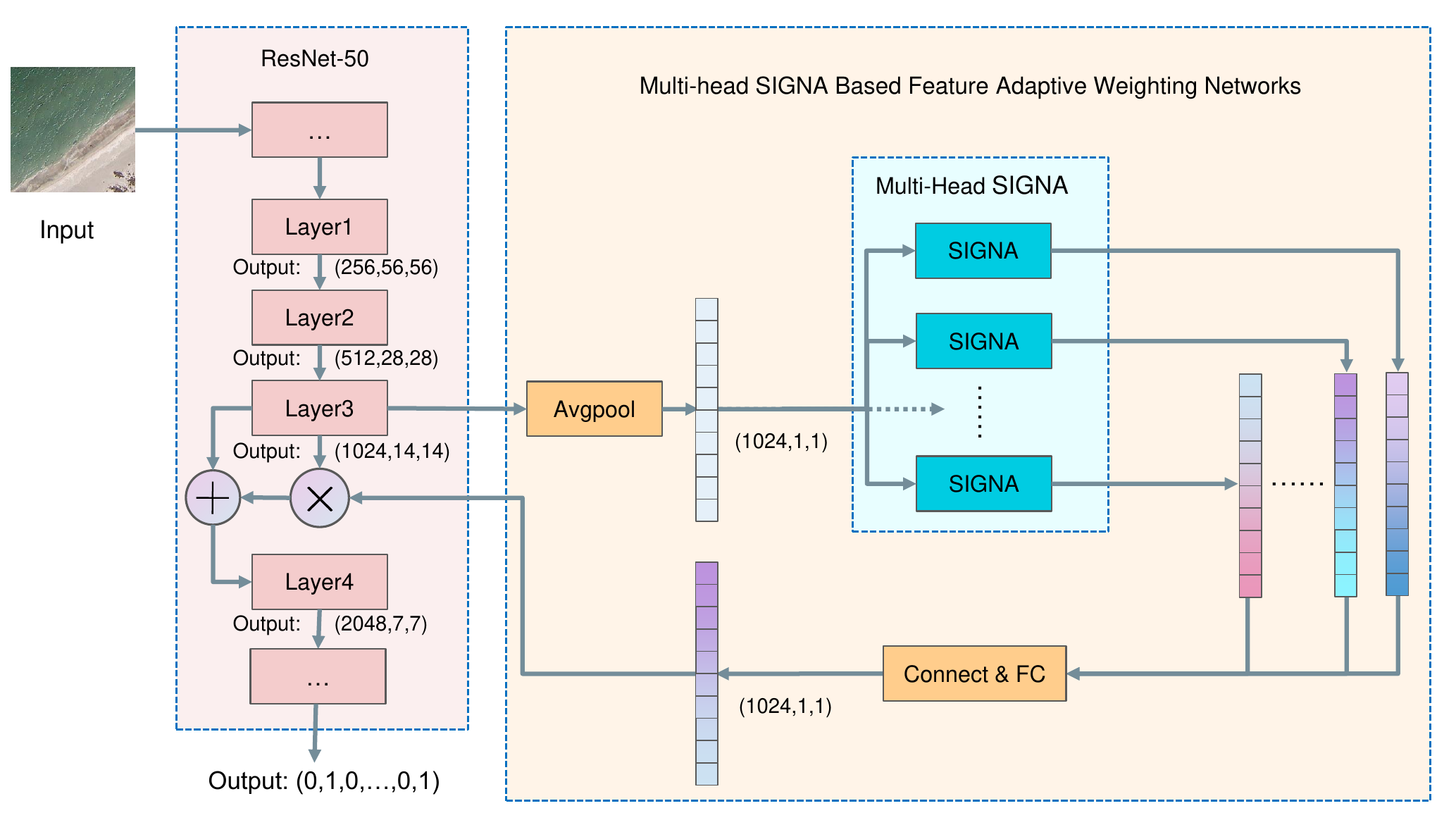}
	\caption{Pineline of Multi-head SIGNA Based Feature Adaptive Weighting Networks: taking ResNet-50 as an example, the workflow of ResNet-50 is that the input is a remote sensing scene image, which is output as a multi label classification result encoded by one hot after passing through several layers of CNN. SIGNA can be inserted into any layer of CNN for function. In this figure, we show the process of inserting layer3 of ResNet-50 for function.}
	\label{fig:pipeline}
\end{figure*}

\subsection{Semantic Interleaving Global Channel Attention}
\label{subsub:semantic interleaving Global Channel Attention}

The following details how to apply semantic interleaving encoding to image channel features and trigger global attention in the implicit semantic feature space.

As shown in Fig.\ref{fig:Attention}(b),
given an image feature $ i \in \mathbb{R}^{  D \times w \times h}$ , use Avgpool to compress it in the spatial dimension. And the squeezed channel feature $z \in \mathbb{R}^{D \times 1 \times 1} $ is obtained.
where $D$ represents the number of channels of the image feature, and $w$ and $h$ represent the width and height of the image feature.
After using semantic interleaving encoding to encode squeezed channel features, the semantic interleaving matrix $M_s$ is obtained. 
A fully connected layer is used to map the squeezed channel features $Z$ to a vector $Z_v$ of the same shape.
The weighting vector $Z_w$ is obtained by matrix multiplication of $M_s$ and $Z_v$.
The value of the n-th row of $Z_w$ is obtained by multiplying the n-th row of $M_s$ by $Z_v$, which means that in the semantic feature space the first value of $Z$ is multiplied by each value of $Z$.
In this way, the similarity of this eigenvalue with other eigenvalues in the semantic feature space is obtained.
Finally, this vector can adaptively weight the feature maps of the full channel.

The same is the channel attention, the attention mechanism of SIGNA and SE-Net\cite{hu2018squeeze} is different. SE-Net models the importance of each channel through two fully-connected layers, without involving label correlation.(Fig.\ref{fig:Attention}(a))
And SIGNA is to achieve global attention in the semantic feature space with label correlation. It calculates the similarity between each channel and other channels in the semantic feature space, and obtains the importance of each channel.(Fig.\ref{fig:Attention}(b))

\subsection{Multi-head SIGNA Based Feature Adaptive Weighting Networks}
\label{subsub:Multi-head SIGNA Based Feature Adaptive Weighting Networks}

In order to better utilize the characteristics of remote sensing image data sets, SIGNA is designed as a plug-and-play network: Multi-head SIGNA Based Feature Adaptive Weighting Networks.
On the basis of SIGNA, this network introduces the idea of multi-head mechanism and residual structure. It can be inserted into any layer of the CNN backbone to play a role.

Fig.\ref{fig:pipeline} shows the pipeline for inserting the proposed network into any layer of the CNN.
First, the squeezed channel features are obtained through a pooling operation with a global receptive field, and each feature map is compressed into a feature value.
Next, connect $N$ SIGNA modules to produce $N$ weighting vectors. $N$ weighting vectors are concatenated in the same dimension, and then map into a weighting vector through a fully connected layer.
This final weighting vector weights each feature map through a broadcast mechanism, enhancing or suppressing feature maps that are related to the label or not related to the label. 
In order to prevent the weighting effect from being too much, inspired by the idea of residual structure, an original feature map is added after the weighted feature map.
In this way, the original visual features can be preserved when the semantic relationship constraint effect is not good.

\section{Experiments and Analysis}
\label{sec:Experiments and Analysis}

To verify the effectiveness of the proposed method, we run it on three publicly available multi-label remote sensing image data sets. 
In this section, we introduce these data sets and some experimental details, then draw some useful conclusions from the analysis.

\subsection{Data Sets}
\label{subsec:Data Sets}

\begin{figure}[t]
	\centering
	\includegraphics[width=1.0\linewidth]{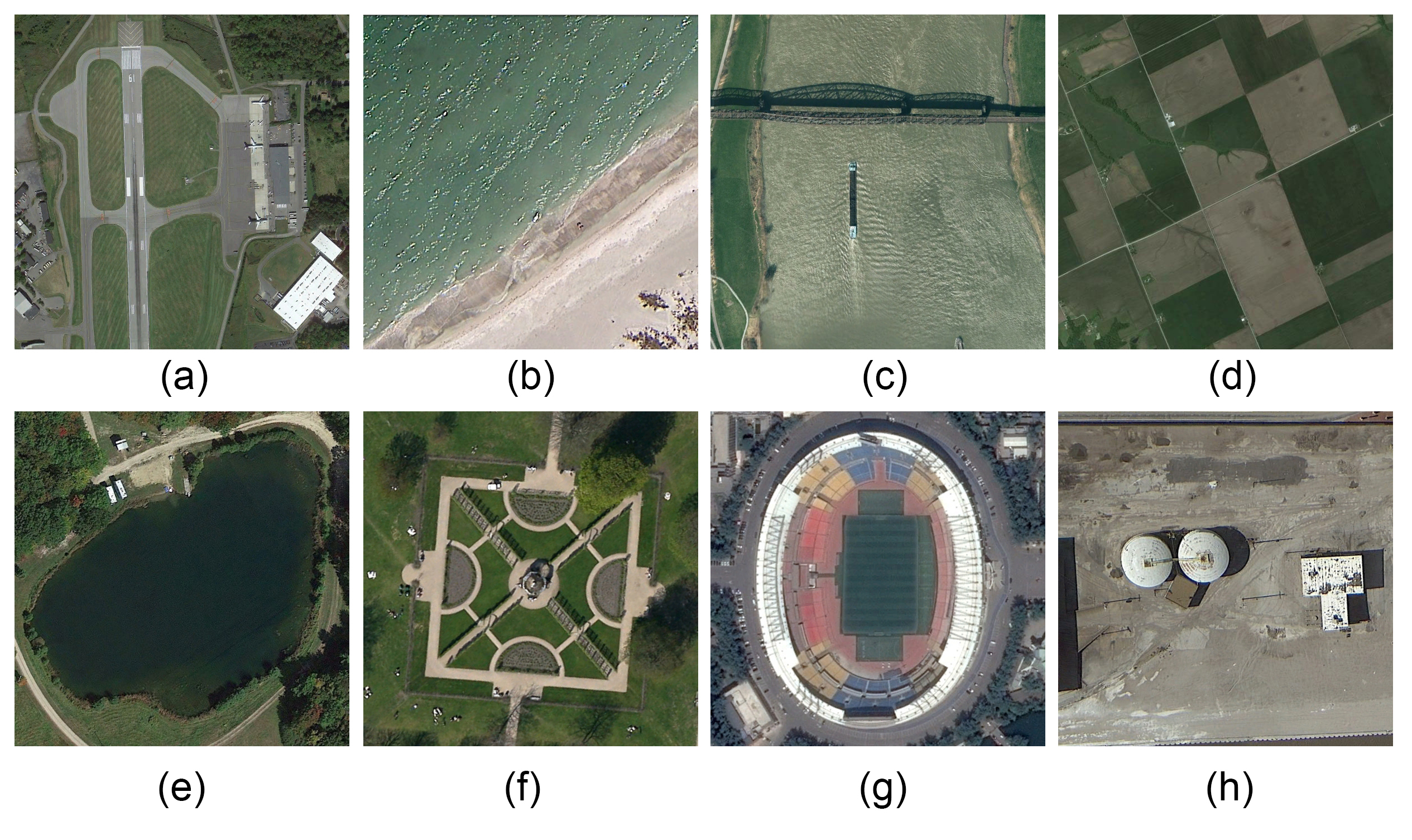}
	\caption{Examples of the AID multi-label data set. (a) Airplane, buildings, cars, grass, pavement, trees. (b) Chaparral, sand, sea.  (c) Grass, pavement, ship, water. (d) Soil, buildings, fields, trees.  (e) Soil, buildings, grass, mobile-home, pavement, trees, water. (f) Buildings, cars, grass, pavement, trees. (g) Buildings, cars, courts, docks, grass, pavement, trees, water. (h) buildings, pavement, tanks.}
	\label{fig:aid}
\end{figure}

\begin{figure}[t]
	\centering
	\includegraphics[width=1.0\linewidth]{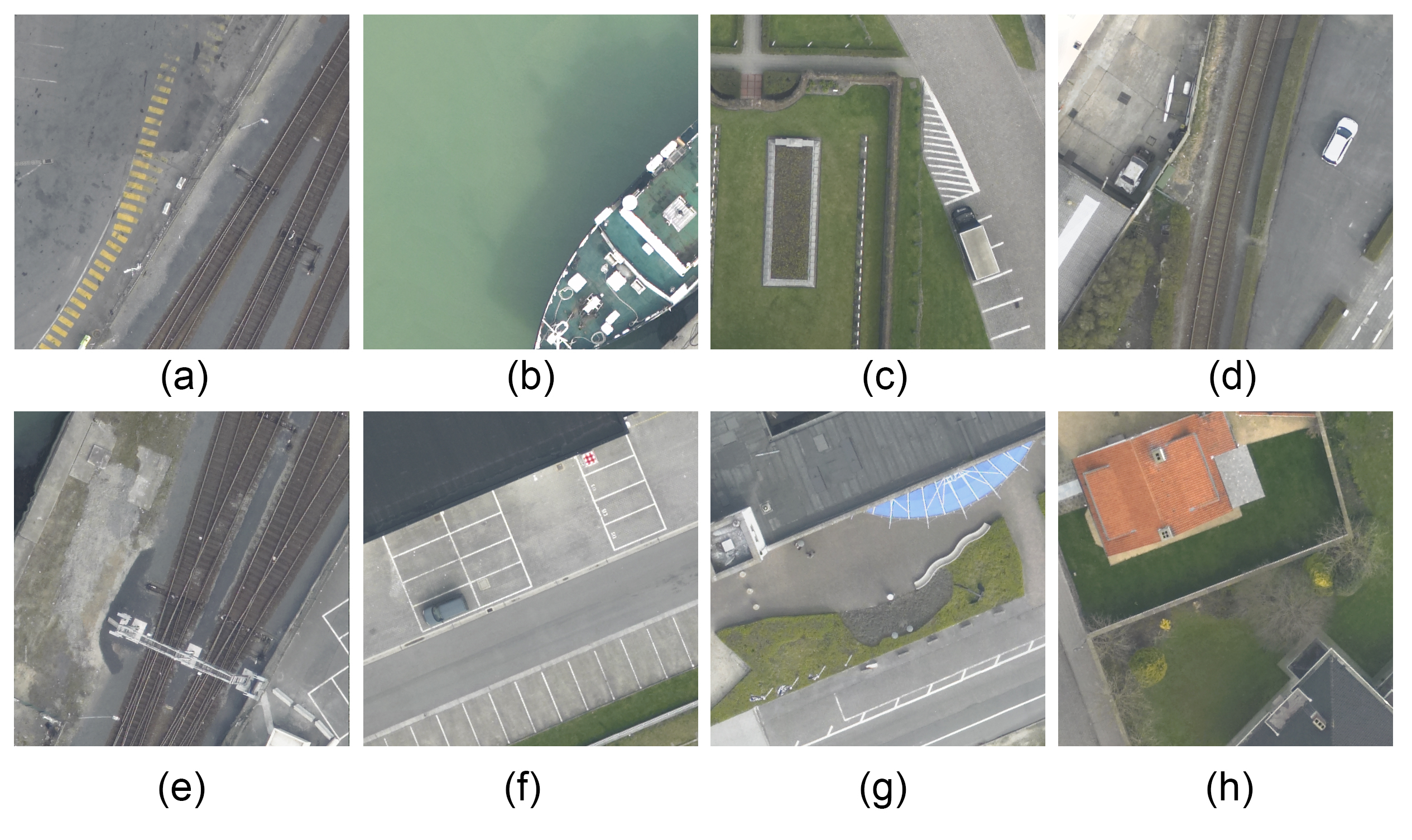}
	\caption{Examples of the UCM multi-label data set. (a) Impervious, clutter, vegetation. (b) Impervious, water, boat. (c) Impervious, vegetation, buildings, cars. (d) Impervious, clutter, vegetation, buildings, trees, cars. (e) Field. (f) Sand, sea. (g) Cars, grass, mobile-home, pavement.  (h) Soil, buildings, cars, pavement, tanks. }
	\label{fig:ucm}
\end{figure}

\begin{figure}[t]
	\centering
	\includegraphics[width=1.0\linewidth]{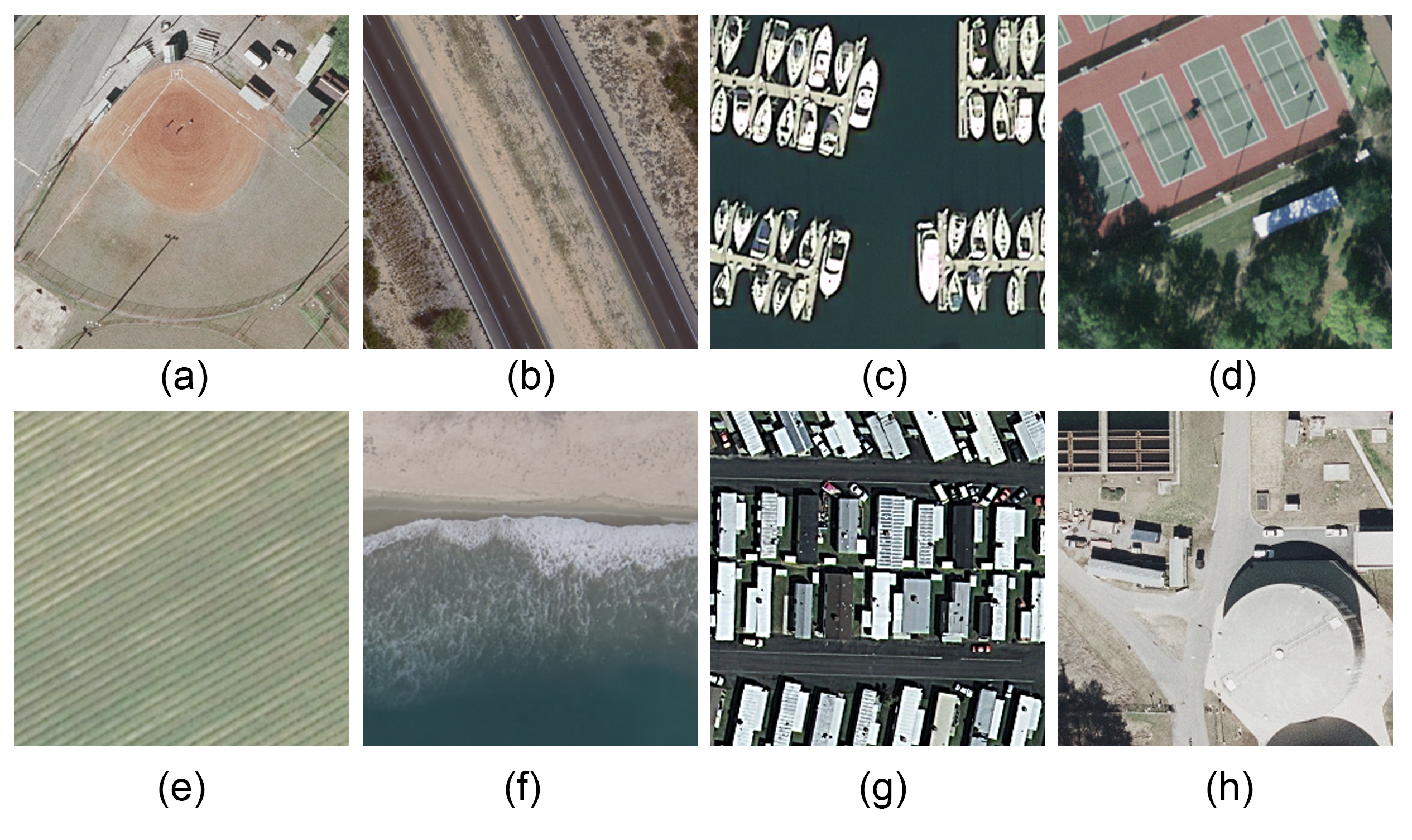}
	\caption{Examples of the DFC15 multi-label data set. (a) Soil, buildings, cars, pavement. (b) Soil, chaparral, pavement, trees.  (c) Dock, ship, water. (d) Court, grass, trees. (e) Impervious, water, clutter, building. (f) Impervious, vegetation, building, car. (g) Impervious, vegetation, building. (h) Impervious, vegetation, building, tree.}
	\label{fig:dfc15}
\end{figure}

We conduct experiments on three multi-label remote sensing data sets: AID, UCM and DFC15 data sets. These three data sets are described in detail below.

\begin{itemize}
\item UCM Multi-label Data Set\cite{chaudhuri2017multilabel}: The UCM multi-label data set is a relabeled version of the UCM data set. The original UCM data set was produced by the United States Geological Survey (USGS) National Map and contains 2100 images grouped into 21 scene categories. After reprocessing the original UCM data set, each image in the current UCM multi-label data set is 256×256 pixels and contains a total of 17 classes: airplane, bare soil, buildings, cars, chaparral, court, dock, field, grass, mobile home, pavement, sand, sea, ship, tanks, trees, and water. Fig. \ref{fig:ucm} displays some multilabel examples.

\item AID Multi-label Data Set\cite{hua2019label}: This data set is recreated from the widely used AID data set. AID data set is a public remote sensing data set manually intercepted by Google Earth and released by Wuhan University, which is specially used for aerial image classification. The AID multi-label data set selects 100 images of each scene category from the original data set, for a total of 3000 multilabel images divided into 30 scene categories with an image size of 600 × 600 pixels. Similarly, the newly defined labels of the AID data set also contain 17 categories, which are consistent with the labels of the UCM data set mentioned above. Fig. \ref{fig:aid} illustrates some multilabel examples from this data set.

\item DFC15 Multilabel Data Set\cite{hua2019label}: This data set is created from the well-known DFC15 data set. It has a total of 3342 images, each of which is 256 × 256 pixels, and contains a total of 8 label categories: building, boat, car, clutter, impervious, water, vegetation and trees. 
Some examples of multilabel annotations are given in Fig. \ref{fig:dfc15}.

\end{itemize}

\begin{figure*}
    \centering

    \begin{minipage}[b]{1\textwidth}
    \includegraphics[width=0.33\textwidth]{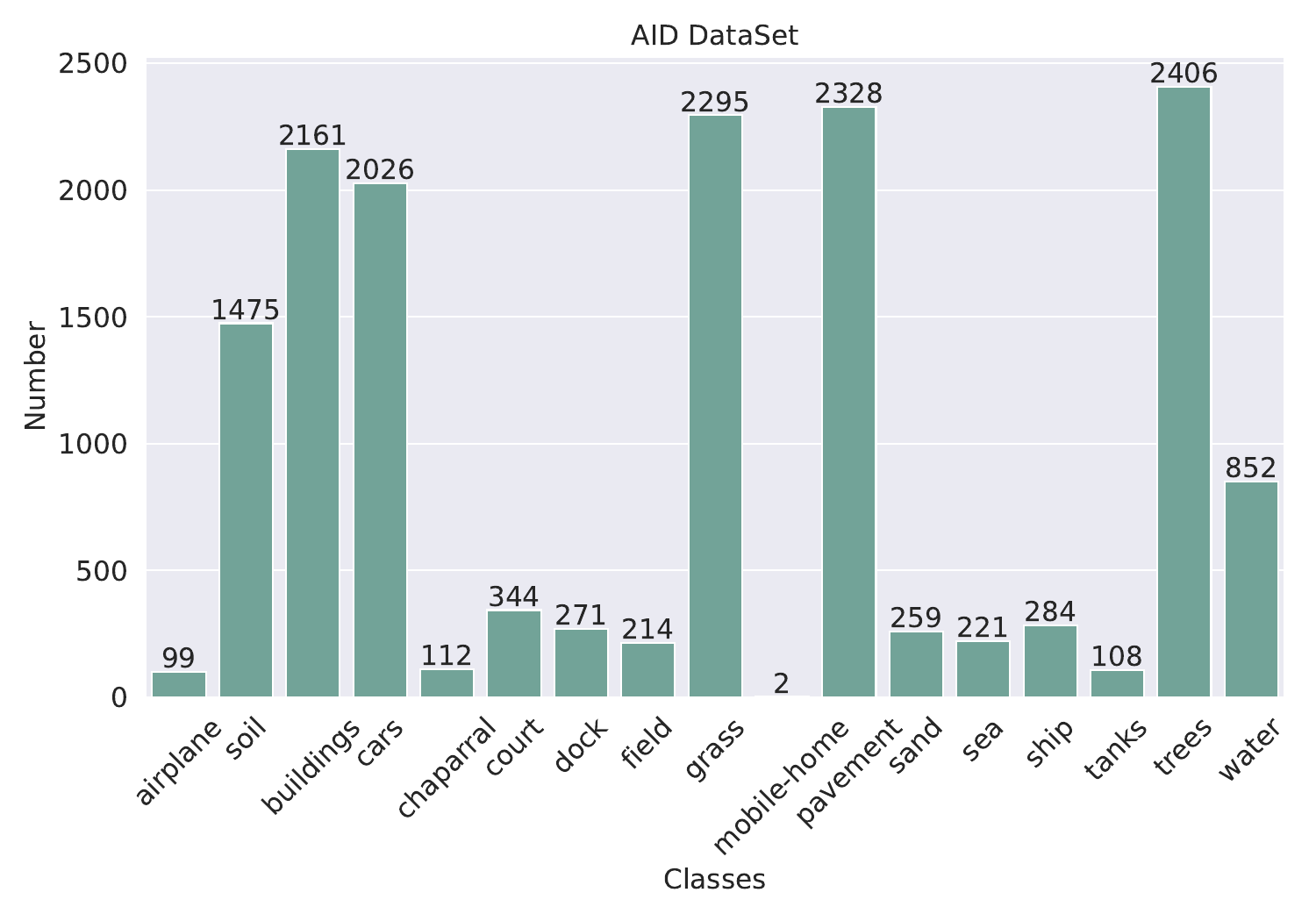} 
    \includegraphics[width=0.33\textwidth]{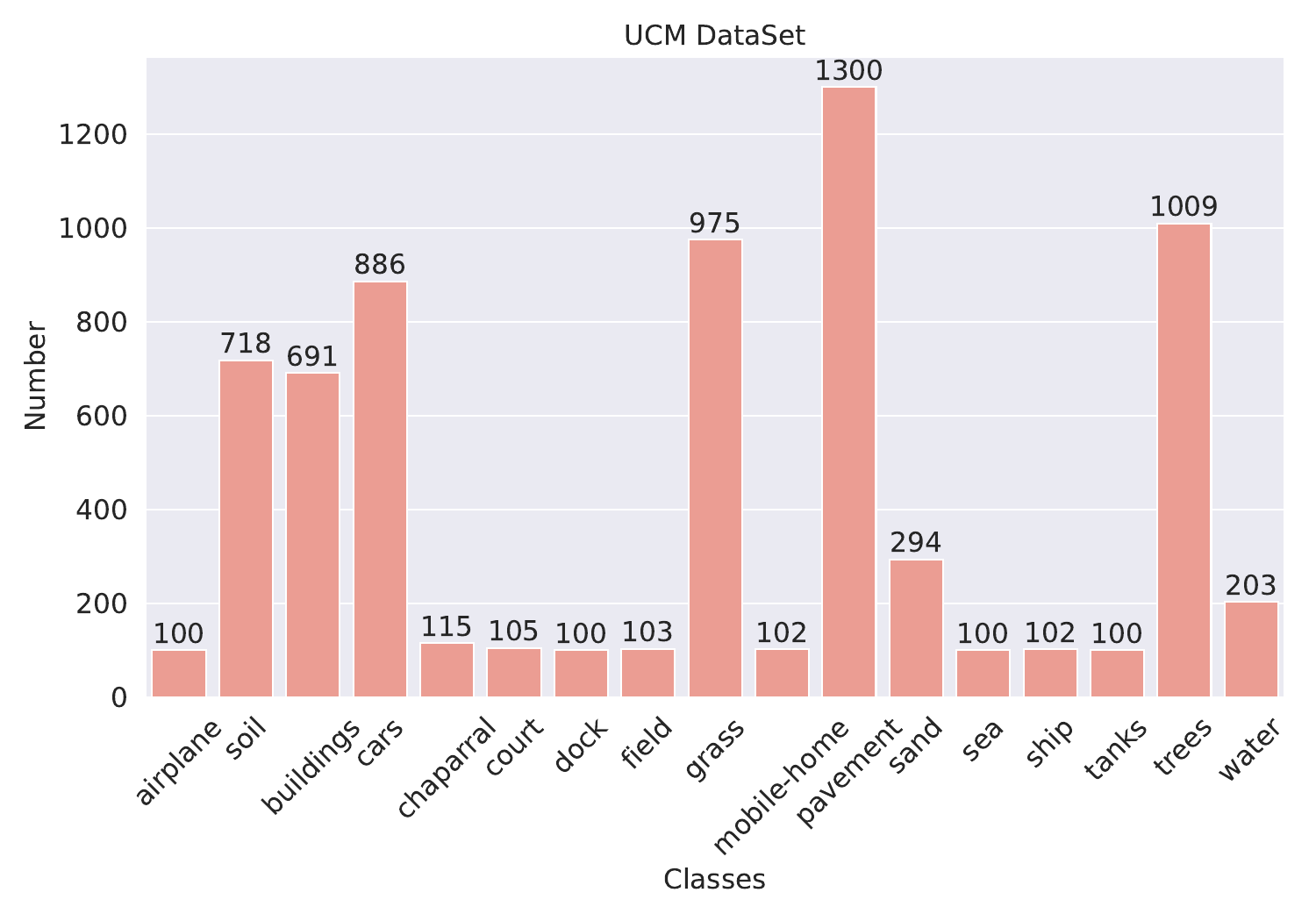}
    \includegraphics[width=0.33\textwidth]{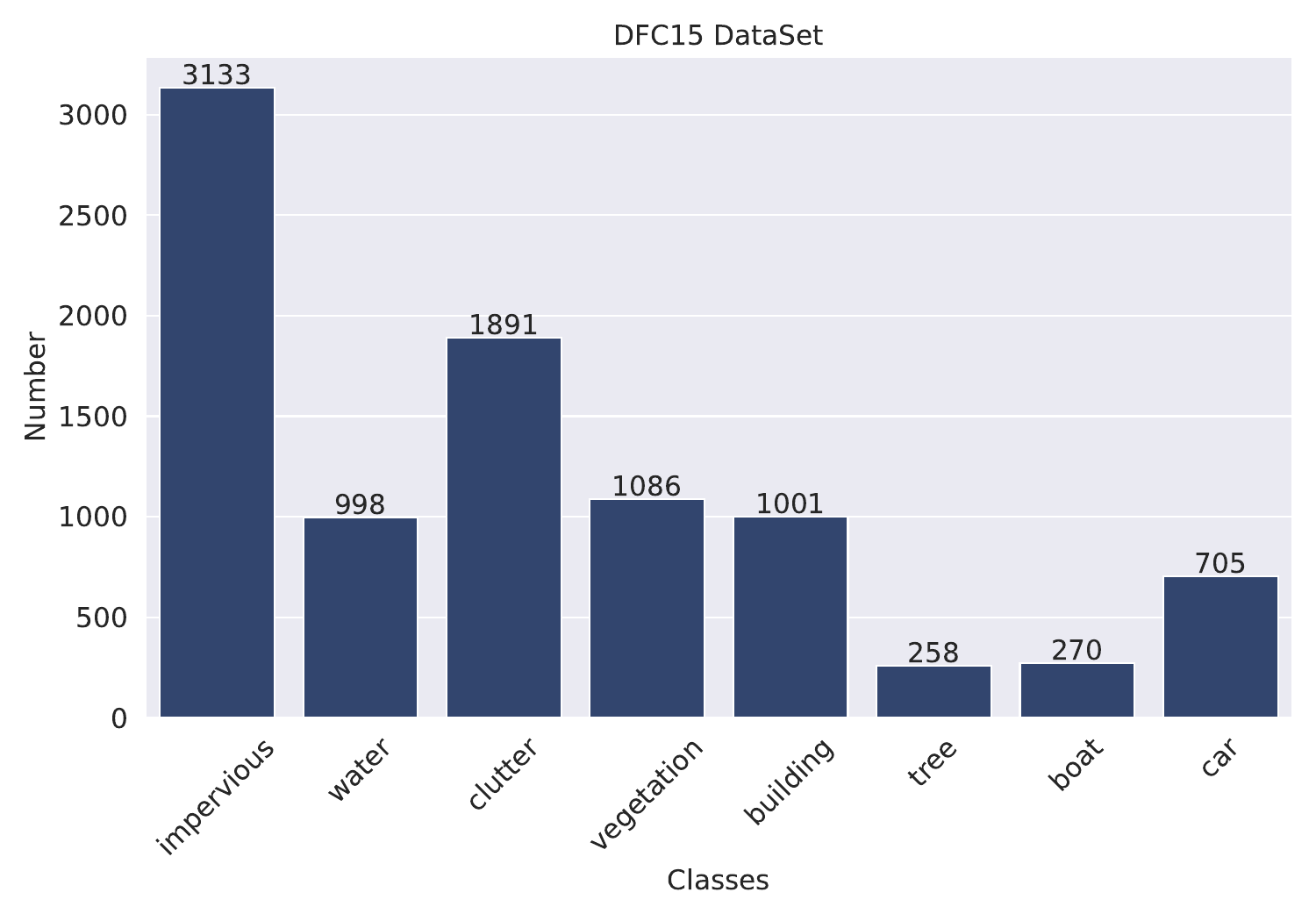}
    \end{minipage}
    \caption{Label statistics for AID data set, UCM data set and DFC15 data set}

    \label{fig:label statistics}
\end{figure*}

\begin{figure*}
    \centering

    \begin{minipage}[b]{1\textwidth}
    \includegraphics[width=0.33\textwidth]{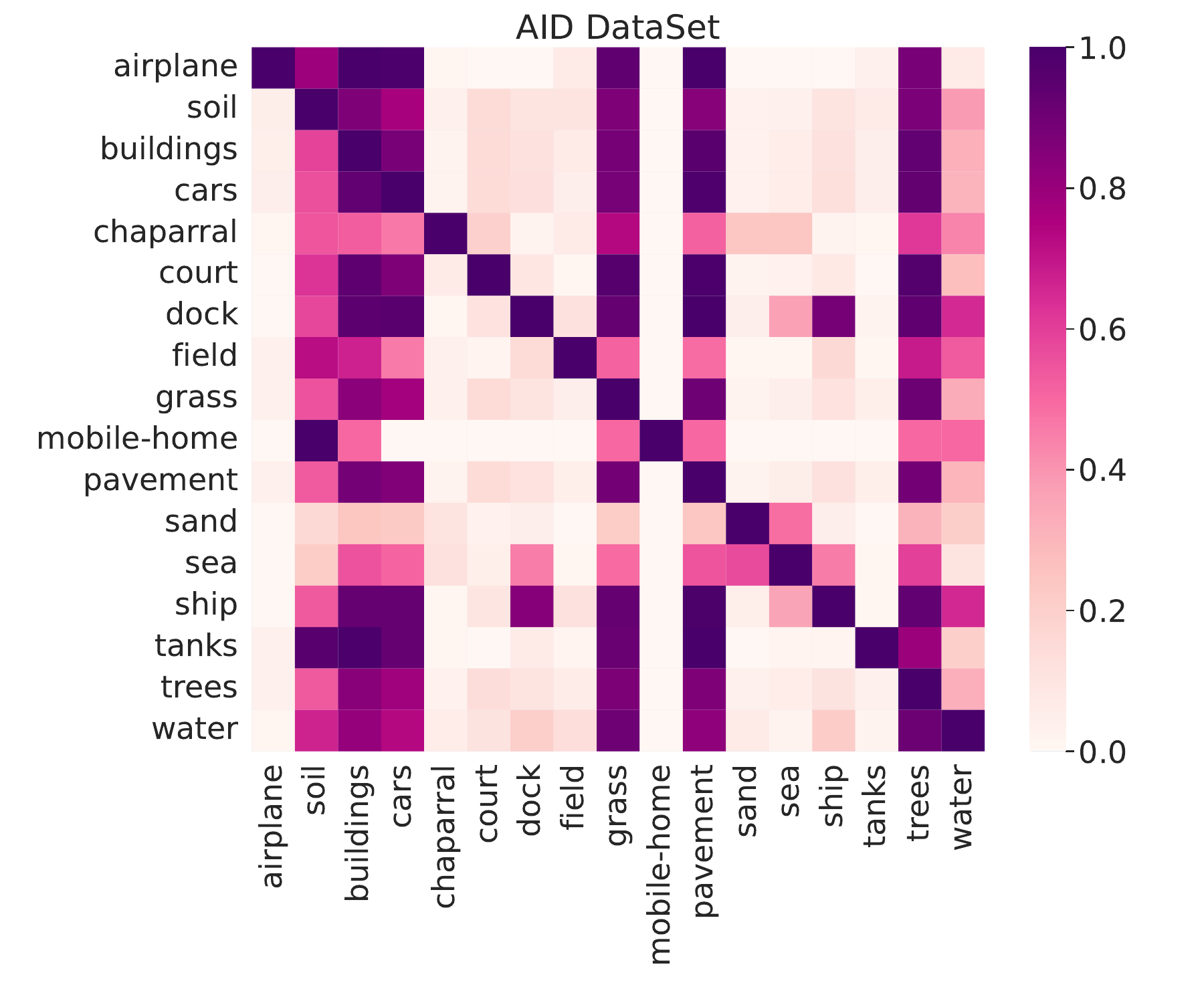} 
    \includegraphics[width=0.33\textwidth]{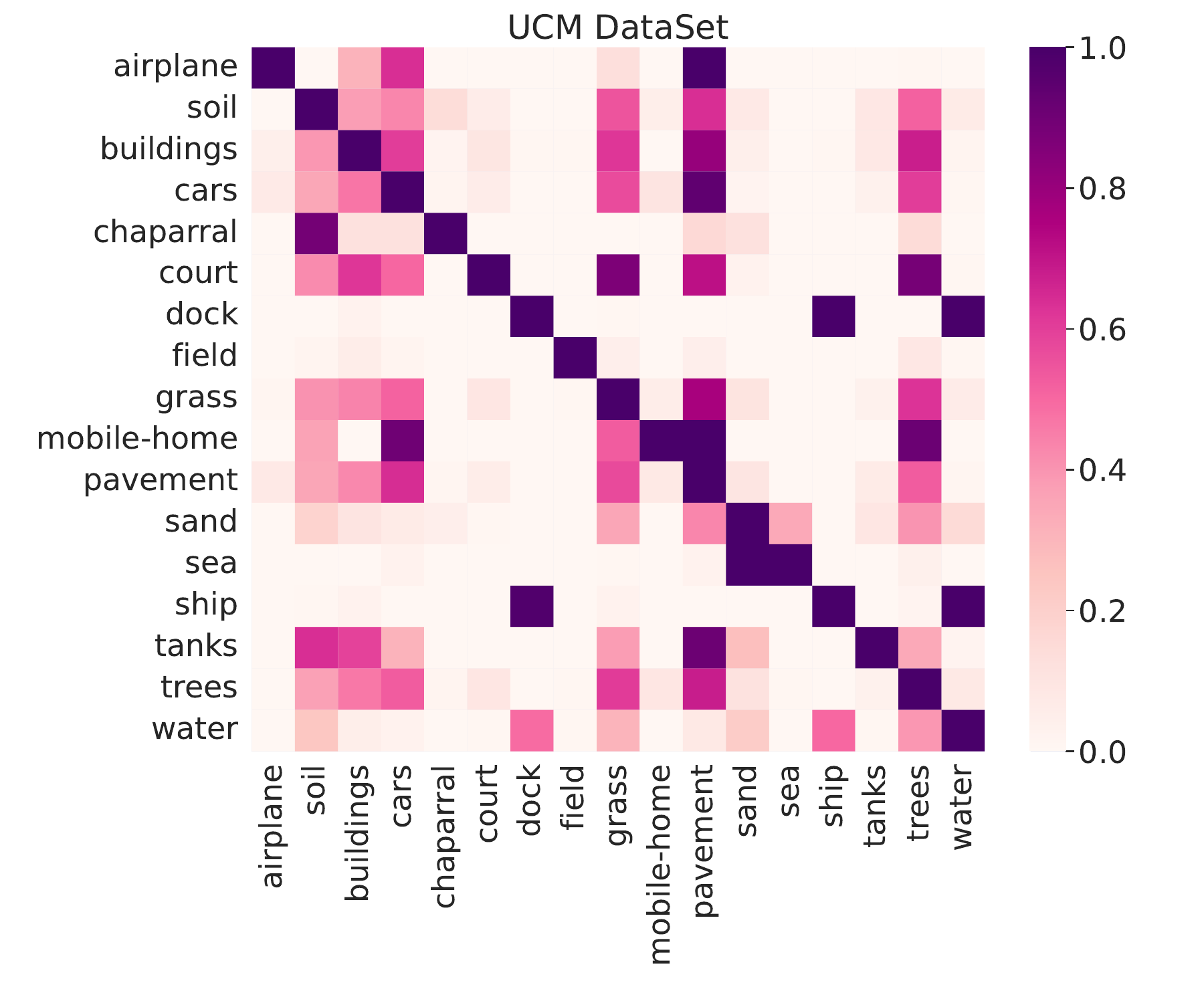}
    \includegraphics[width=0.33\textwidth]{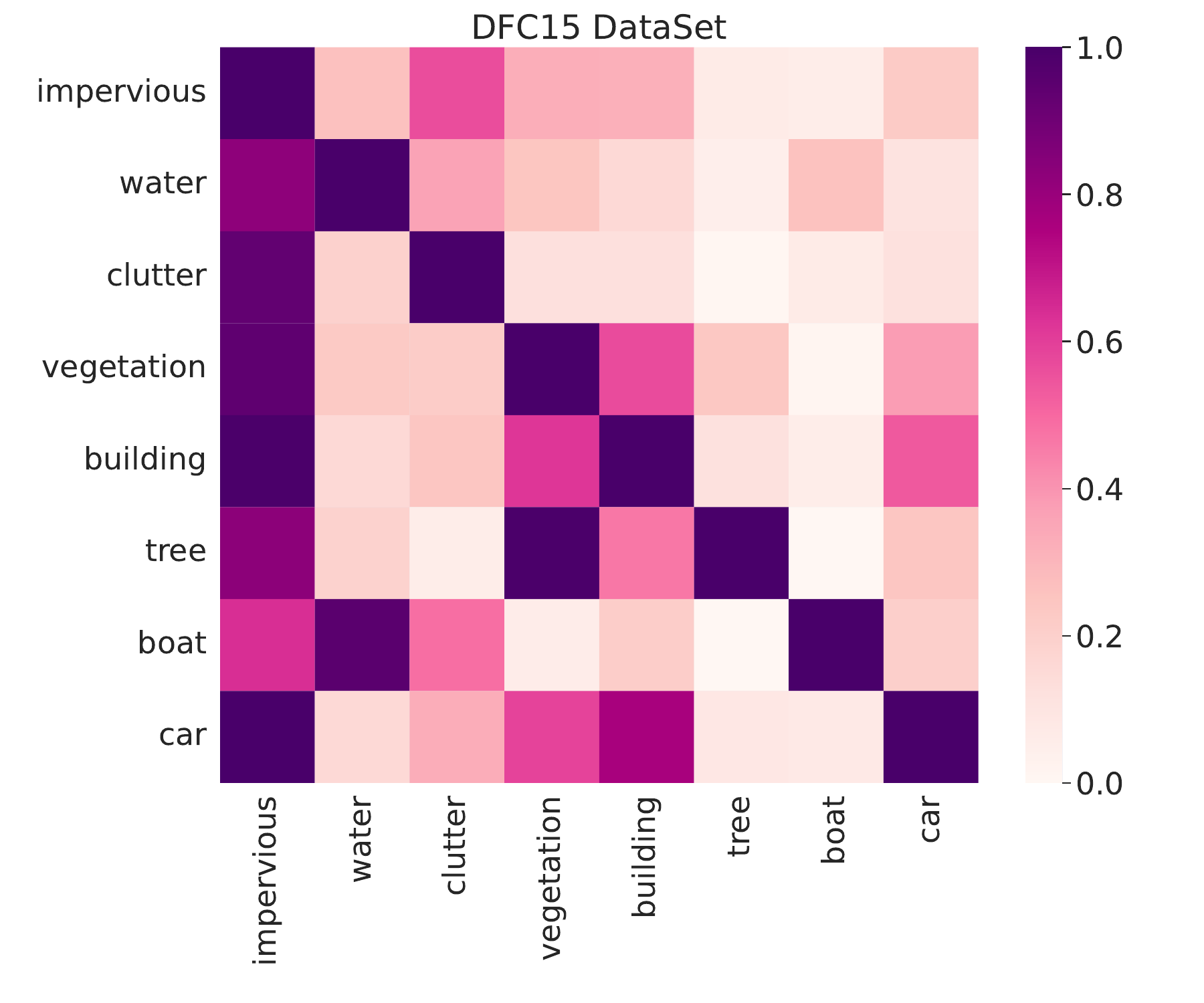}
    \end{minipage}
    \caption{Label co-occurrence matrix for AID data set, UCM data set and DFC15 data set}

    \label{fig:label matrix}
\end{figure*}

The numbers of images in each class in the three data sets are shown in Fig. \ref{fig:label statistics}.
Fig. \ref{fig:label matrix} displays the label co-occurrence matrix in three data sets.
From the label statistics, it can be known that the labels of the three data sets are unbalanced, and the number of label samples varies significantly.

The AID and UCM data sets use the same tag set, so some conclusions can be drawn by comparing the two data sets. The simpler the co-occurrence graph, the clearer and more effective the constraints between the labels. Compared with AID data set, the co-occurrence graph of UCM data set has fewer and more obvious edges. This indicates that the co-occurrence relationship of AID data set is complex and fragmentary, while the co-occurrence relationship of UCM data set is more obvious. Therefore, it can be speculated that the label relation method may work better on UCM data set than on AID data set. 
experimental data in Table \ref{tab:UCM-per} and \ref{tab:AID-per}, and the section \ref{subsub:Per-Class Case Studies} demonstrate this point.

\begin{figure*}
    \centering

    \begin{minipage}[b]{1\textwidth}
    \includegraphics[width=0.33\textwidth]{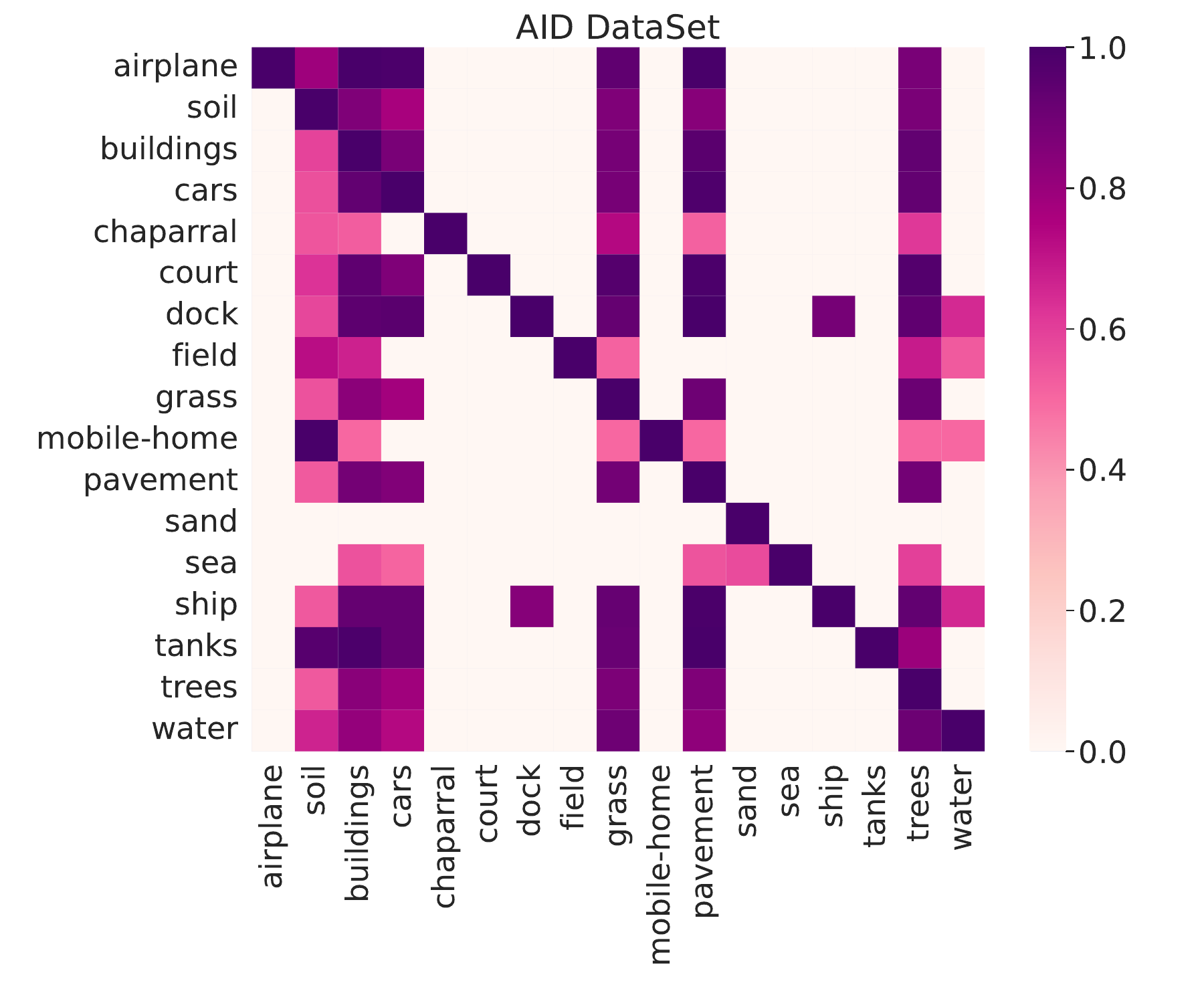} 
    \includegraphics[width=0.33\textwidth]{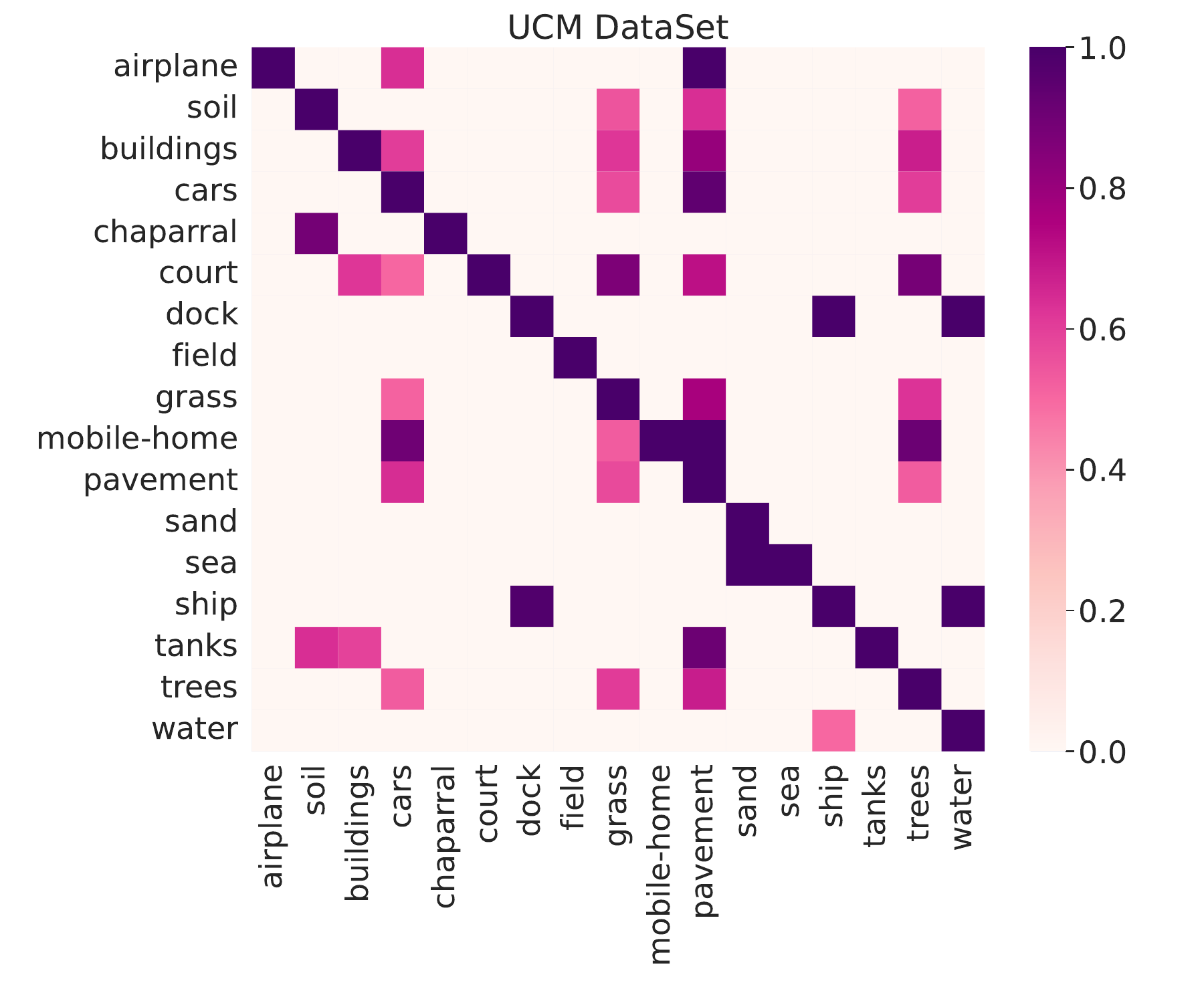}
    \includegraphics[width=0.33\textwidth]{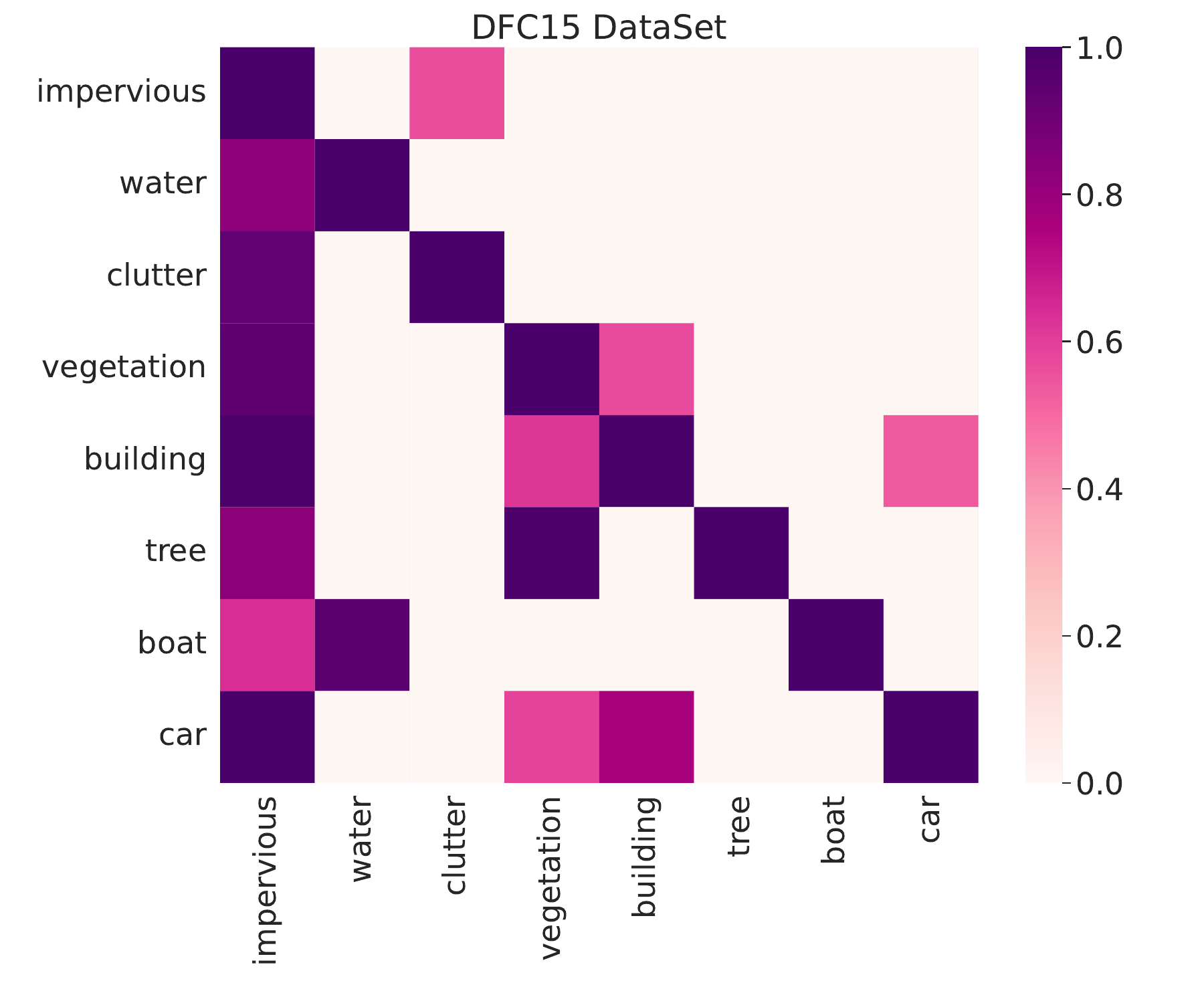}
    \end{minipage}
    \caption{Label co-occurrence matrix for AID data set, UCM data set and DFC15 data set after formula \ref{eq:smoothing}.}

    \label{fig:label matrix2}
\end{figure*}

\subsection{Experimental setup}
\label{subsec:Experimental setup}

\subsubsection{Evaluation Metrics}
\label{subsub:Evaluation Metrics}

In our experiments, the performance of different models is evaluated quantitatively using the example-based F1 and F2 scores\cite{zhang2013review} as evaluation metrics. To perform a comprehensive evaluation of the network, the classification performance of each category is evaluated using label-based F1 and F2, thus analyzing the impact of the network on each label. 

Specifically, the example-based F1 and F2 scores are calculated as follows.

\begin{equation}\label{eq:F1F2e}
F\beta_e=\left(1+\beta^{2}\right) \frac{P_{e} R_{e}}{\beta^{2} P_{e}+R_{e}}, \quad \beta=1,2
\end{equation}
\begin{equation}\label{eq:PR}
P_{e}=\frac{1}{n} \sum_{i=1}^{n}\frac{\mathrm{TP}_{e}}{\mathrm{TP}_{e}+\mathrm{FP}_{e}}, \quad R_{e}=\frac{1}{n} \sum_{i=1}^{n}\frac{\mathrm{TP}_{e}}{\mathrm{TP}_{e}+\mathrm{FN}_{e}}
\end{equation}
where $P_e$ and $R_e$ are the example-based precision and recall, respectively. ${TP}_e$ represents the number of correctly predicted positive labels in an example. ${FP}_e$ represents the number of unrecognized positive labels in an example. ${FN}_e$ represents the number of mispredicted negative labels in an example. $n$ represents the number of data set images.

The label-based F1 and F2 scores are calculated as follows.

\begin{equation}\label{eq:F1F2l}
F\beta_l=\left(1+\beta^{2}\right) \frac{P_{l} R_{l}}{\beta^{2} P_{l}+R_{l}}, \quad \beta=1,2
\end{equation}

\begin{equation}\label{eq:PR2}
P_{l}=\frac{1}{c} \sum_{j=1}^{c}\frac{\mathrm{TP}_{l}}{\mathrm{TP}_{l}+\mathrm{FP}_{l}}, \quad R_{l}=\frac{1}{c} \sum_{j=1}^{C}\frac{\mathrm{TP}_{l}}{\mathrm{TP}_{l}+\mathrm{FN}_{l}}
\end{equation}
where $P_l$ and $R_l$ are the label-based precision and recall, respectively. ${TP}_l$ represents the number of all correctly predicted positive labels in a class. ${FP}_l$ represents the number of unrecognized positive labels in a class. ${FN}_l$ represents the number of negative labels for all mispredictions in a class. $c$ represents the number of data set classes.

\subsubsection{Loss Functions and Multi-classification}
\label{subsub:Loss Functions and Multi-classification}
We connect a sigmoid activation function at the end of the network, and treat each value of the output layer as a binomial distribution. So that the multi-label problem can be successfully transformed into a binomial classification problem on each label. The calculation formula of the sigmoid activation function is as follows.
\begin{equation}\label{eq:Loss}
sigmoid(x)=\frac{1}{1+\exp (-x)} \in(0,1).
\end{equation}

 We use multi-label binary cross-entropy loss (BCE) function as the loss function and is calculated as follows.
\begin{equation}\label{eq:Loss2}
Loss=-\sum_{i=1}^{n} \sum_{j=1}^{c} y_{i j} \log \hat{y}_{i j}+\left(1-y_{i j}\right) \log \left(1-\hat{y}_{i j}\right)
\end{equation}
where $c$ is the number of labels and $n$ is the number of examples.

\subsubsection{Training details}
\label{subsub:Training details}
We used three data enhancement strategies in the training process: Random Horizontal Flip, Random Vertical Flip and Color Jitter (brightness = contrast = saturation = (0.6 , 1.4)).
The model is trained using the Adam optimizer, and the initial weights of the CNN backbone are set to the weights of the CNN model pre-trained on the generic data set. The initial learning rate is set to 0.001. Every 25 epochs decays by 0.1. The batch size is set to 16, and we train the model for 80 epochs.

We randomly selected 70\% of the samples from each of the three data sets as training set, 10\% as validation set, and finally 20\% as test set for training and testing respectively. 
The GNN in SIGNA is set to two layers, and the output dimensions are the number of channels/2 and the number of channels respectively;
The input to the GNN is the 300-dimensional glove word embedding vector for each label.
All experiments are based on the same hardware and software conditions as follows: GPU: GeForce GTX 3090; OS: Ubuntu 18.04.3 LTS; CUDA Version: 11.1.0; PyTorch Version: 1.9.1 for cu111; Torch-Geometric 2.0.3 for cu111 and TorchVision Version: 0.10.1 for cu111. 

\subsection{Comparison of Different Settings}
\label{subsec:Comparison of Different Settings}

As shown in section \ref{sec:Proposed Approach}, the SIGNA networks can select any layer of the backbone network to insert for global channel attention, its GNN can use the currently popular GCN, GraphSAGE, or GAT for feature extraction, and SIGNA can choose the number of different headers to use the multi-head mechanism.
Below we will conduct experiments on the backbone networks ResNet-18 and ResNet-50 to discuss these combination options and try to find the best combination.

\begin{table}[htbp]
  \centering
  \caption{Compare the number of multiheads in the SIGNA networks.}
    \begin{tabular}{|c|c|c|c|c|}
    \hline
    \hline
          & \multirow{2}[4]{*}{Head} & UCM   & AID   & DFC15 \bigstrut\\
\cline{3-5}          &       & F1    & F1    & F1 \bigstrut\\
    \hline
    \multirow{5}[2]{*}{ResNet-18} & 1     & \textbf{93.00 } & 92.18  & 96.81  \bigstrut[t]\\
          & 2     & 92.85  & 92.38  & 96.63  \\
          & 4     & 92.87  & 92.41  & 96.81  \\
          & 6     & 92.93  & \textbf{92.71 } & \textbf{96.96 } \\
          & 8     & 92.37  & 92.31  & 96.88  \bigstrut[b]\\
    \hline
    \multirow{5}[2]{*}{ResNet-50} & 1     & 93.78  & 92.81  & 97.33  \bigstrut[t]\\
          & 2     & 93.76  & 92.69  & 97.35  \\
          & 4     & 93.84  & 92.83  & 97.24  \\
          & 6     & \textbf{93.92 } & \textbf{92.93 } & \textbf{97.37 } \\
          & 8     & 93.87  & 92.88  & 97.34  \bigstrut[b]\\
    \hline
    \hline
    \end{tabular}%
  \label{tab:compare_heads}%
\end{table}%

\subsubsection{Performance with different number of heads}
\label{subsub:Performance with different number of heads}

We conducted experiments using 1, 2, 4, 6, and 8 heads, each representing one SIGNA module, with channel features replicated N times into N SIGNA module. 
As shown in Table \ref{tab:compare_heads}, there is a little difference in performance for choosing different number of heads. 
Except for ResNet-18 which performs better for single-head SIGNA on UCM data set, the performance of multi-head SIGNA is better than that of single-head SIGNA in all other experiments. 
The F1 scores of 6-heads SIGNA module are optimal for 5 times in two backbone networks and three data sets, so the multi-heads number is set as 6.

\subsubsection{Selecting the performance of the different layers of CNN}
\label{subsub:Selecting the performance of the different layers of CNN}

SIGNA can select any layer of the backbone network for channel attention aggregation. We choose layer1, layer2, layer3 and layer4 of ResNet-18 and ResNet-50 for our experiments. The output shape of each layer of ResNet-18 is (64, 64, 64), (128, 32, 32), (256, 16, 16), (512, 8, 8). And the output shape feature maps of each layer of ResNet-50 are (256, 56, 56), (512, 28, 28), (1024, 14, 14), (2048, 7, 7).
As shown in the Talbe \ref{tab:compare_layer}, AID\&ResNet-18, DFC15\&ResNet-18, and UCM\&ResNet-50 in the experiment achieved the best performance when SIGNA was inserted in the second layer; UCM\&ResNet-18 and DFC15\&ResNet-50 achieved the best performance when inserted in the first layer. This indicates that SIGNA may work better in the shallow layer of CNN.
This may be because: in the shallow layer of the backbone network, the number of channels is small, and the size of the feature map is large, which mainly contains the local detailed features of the image. In the deep layer of the backbone network, the number of channels is large, and the size of the feature map is small, which mainly contains abstract semantic information.
Many labels of remote sensing images are shallow information, such as ocean texture, field texture, etc.
And there are many labels of small-sized objects in remote sensing images, such as cars, so larger feature maps work better for the proposed SIGNA.

\begin{table}[htbp]
  \centering
  \caption{Compare the performance of selecting different layers of CNN.}
    \begin{tabular}{|c|c|c|c|c|}
    \hline
    \hline
    \multirow{2}[4]{*}{} & \multirow{2}[4]{*}{Layer} & UCM   & AID   & DFC15 \bigstrut\\
\cline{3-5}          &       & F1    & F1    & F1 \bigstrut\\
    \hline
    \multirow{4}[2]{*}{ResNet-18} & 1     & \textbf{92.93 } & 92.51  & 96.92  \bigstrut[t]\\
          & 2     & 92.93  & \textbf{92.71 } & \textbf{97.12 } \\
          & 3     & 92.85  & 91.57  & 97.05  \\
          & 4     & 92.60  & 92.69  & 96.95  \bigstrut[b]\\
    \hline
    \multirow{4}[2]{*}{ResNet-50} & 1     & 93.76  & 92.80  & \textbf{97.59 } \bigstrut[t]\\
          & 2     & \textbf{93.92 } & 92.93  & 97.57  \\
          & 3     & 93.67  & \textbf{92.97 } & 97.39  \\
          & 4     & 93.86  & 92.88  & 97.46  \bigstrut[b]\\
    \hline
    \hline
    \end{tabular}%
  \label{tab:compare_layer}%
\end{table}%

\subsubsection{Performance using different GNN }
\label{subsub:Performance using different GNN }

In the semantic Interleaving encoder, we choose the current widely used GCN, GraphSAGE and GAT with excellent performance as relational feature extractors for our experiments. As shown in Table \ref{tab:compare_GNN}, in the experiments, UCM\&ResNet-18, DFC15\&ResNet-18, UCM\&ResNet-50, and AID\&ResNet-50, a total of four combinations, achieve the best performance when GraphSAGE is used in SIGNA. Therefore, we choose GraphSAGE for extracting relational features in the semantic Interleaving encoder.

\begin{table}[htbp]
  \centering
  \caption{Compare the performance of using different GNNs.}
    \begin{tabular}{|c|c|c|c|c|}
    \hline
    \hline
    \multirow{2}[4]{*}{} & GNN   & UCM   & AID   & DFC15 \bigstrut\\
\cline{3-5}          & Type  & F1    & F1    & F1 \bigstrut\\
    \hline
    \multirow{3}[2]{*}{ResNet-18} & GCN   & 92.79  & \textbf{92.71 } & 96.96  \bigstrut[t]\\
          & GraphSAGE & \textbf{92.90 } & 92.56  & \textbf{97.12 } \\
          & GAT   & 92.83  & 92.69  & 96.96  \bigstrut[b]\\
    \hline
    \multirow{3}[2]{*}{ResNet-50} & GCN   & 93.92  & 92.93  & \textbf{97.57 } \bigstrut[t]\\
          & GraphSAGE & \textbf{93.96 } & \textbf{92.96 } & 97.46  \\
          & GAT   & 93.91  & 92.70  & 97.40  \bigstrut[b]\\
    \hline
    \hline
    \end{tabular}%
  \label{tab:compare_GNN}%
\end{table}%

\subsection{Comparisons With Other Methods And Baselines}
\label{subsec:Comparisons With Other Methods And Baselines}
For comprehensive evaluation, we compare the proposed method with the following state-of-the-art multi-label classification methods.

\begin{table*}[htbp]
  \centering
  \caption{F1, P and R scores of our proposed method and baseline on each category of the UCM data set.}
    \resizebox{.98\textwidth}{!}{
\begin{tabular}{|c|c|cccc|cccc|cccc|}
    \hline
    \hline
    \multirow{2}[4]{*}{Backbone} & \multirow{2}[4]{*}{Method} & \multicolumn{4}{c|}{UCM}      & \multicolumn{4}{c|}{AID}      & \multicolumn{4}{c|}{DFC15} \bigstrut\\
\cline{3-14}          &       & F1    & F2    & P     & R     & F1    & F2    & P     & R     & F1    & F2    & P     & R \bigstrut\\
    \hline
    \multirow{10}[2]{*}{ResNet-50} & \textbf{Ours}  & \textbf{93.96 } & \textbf{93.94 } & \textbf{94.00 } & \textbf{93.93 } & \textbf{92.96 } & \textbf{92.66 } & \textbf{93.46 } & \textbf{92.46 } & \textbf{97.46 } & \textbf{97.07 } & \textbf{98.12 } & \textbf{96.81 } \bigstrut[t]\\
          & Baseline & 81.40  & 81.73  & 80.86  & 81.95  & 87.44  & 86.36  & 89.31  & 85.65  & 93.13  & 91.62  & 95.74  & 90.65  \\
          & GRN   & 92.40  & 92.66  & 91.98  & 92.83  & 91.93  & 91.42  & 92.79  & 91.08  & 96.24  & 96.07  & 96.53  & 95.95  \\
          & SE-NET & 89.43  & 88.54  & 90.94  & 87.96  & 88.91  & 88.05  & 90.38  & 87.48  & 94.42  & 93.74  & 95.56  & 93.30 \\
          & ML-GCN & 90.36  & 90.57  & 90.03  & 90.70  & 89.58  & 89.52  & 89.69  & 89.48  & -     & -     & -     & - \\
          & SR-Net & 88.67  & 89.11  & 87.96  & 89.40  & 89.97  & 90.30  & 89.42  & 90.52  & -     & -     & -     & - \\
          & AL-RN & 87.93  & 87.41  & 88.81  & 87.07  & 89.96  & 89.35  & 91.00  & 88.95  & -     & -     & -     & - \\
          & CA-BiLSTM & 83.11  & 86.56  & 77.94  & 89.02  & 89.01  & 89.00  & 89.03  & 88.99  & -     & -     & -     & - \\
          & ML-CG & 85.42  & 88.08  & 81.34  & 89.94  & -     & -     & -     & -     & 95.88  & 96.01  & 95.68  & 96.09  \\
          & MSGM  & 84.66  & 85.15  & 83.86  & 85.48  & -     & -     & -     & -     & 93.65  & 93.08  & 94.61  & 92.71  \bigstrut[b]\\
    \hline
    \multirow{9}[2]{*}{VGG16} & \textbf{Ours}  & \textbf{92.97 } & \textbf{93.24 } & \textbf{92.51 } & \textbf{93.43 } & \textbf{91.78 } & 91.69  & \textbf{91.93 } & 91.63  & \textbf{96.97 } & \textbf{96.74 } & \textbf{97.37 } & \textbf{96.58 } \bigstrut[t]\\
          & Baseline & 80.65  & 81.63  & 79.06  & 82.30  & 86.86  & 86.54  & 87.41  & 86.32  & 93.51  & 93.46  & 93.61  & 93.42  \\
          & ML-KNN & 87.48  & 87.89  & 86.82  & 88.16  & 84.23  & 84.48  & 83.82  & 84.65  & -     & -     & -     & - \\
          & MLRSSC-CNN-GNN & 87.76  & 88.15  & 87.11  & 88.41  & 90.01  & 90.13  & 89.83  & 90.20  & -     & -     & -     & - \\
          & Gardner & 86.36  & 85.24  & 88.29  & 84.51  & 84.26  & 83.17  & 86.15  & 82.46  & -     & -     & -     & - \\
          & RBFNN & 88.81  & 88.77  & 88.87  & 88.75  & 87.53  & 86.95  & 88.52  & 86.56  & -     & -     & -     & - \\
          & Stivaktakis & 86.29  & 86.98  & 85.16  & 87.45  & 87.80  & 87.87  & 87.69  & 87.92  & -     & -     & -     & - \\
          & CNN-RNN & 77.09  & 78.74  & 74.49  & 79.88  & 84.53  & 84.82  & 84.06  & 85.01  & -     & -     & -     & - \\
          & Huang & 91.74  & 92.48  & 90.54  & 92.98  & 91.45  & \textbf{91.71 } & 91.03  & \textbf{91.88 } & -     & -     & -     & - \bigstrut[b]\\
    \hline
    Inception-ResNet-v2 & Zhu   & 91.70  & 91.67  & 91.75  & 91.65  & 89.06  & 88.67  & 89.72  & 88.41  & -     & -     & -     & - \bigstrut\\
    \hline
    \hline
    \end{tabular}%
    
    }
  \label{tab:compare others}%
\end{table*}%

\begin{itemize}
	\item GRN\cite{kang2020graph}: This method uses GCN to model the relationships between samples, thus guiding the CNN into a more discriminative metric space.
	
	\item SE-NET\cite{hu2018squeeze}: This method adaptively recalibrates channel-wise feature responses by explicitly modelling interdependencies between channels.
	
    \item ML-GCN\cite{chen2019multi}: The method uses GCN for label correlation extraction, which is explicitly used as a classifier for the final output of the CNN.
	
	\item SR-Net\cite{tan2022transformer}: The method locates the semantic attentional regions in the features extracted by a deep CNN and generates a discriminative content-aware category representation.
	
	\item AL-RN\cite{hua2020relation}: The method extracts high-level label-specific features and locates discriminative regions in these features. Finally, an MLP layer is used to generate label relations for final classification.
	
	\item CA-BiLSTM\cite{hua2019recurrently}: This method uses an attention learning layer to capture class-specific features, and uses the bidirectional LSTM network to model the class dependence for final classification.
	
	\item ML-CG\cite{lin2021multilabel}: This method infers label correlations from a specific data set and ConceptNet (a common-sense knowledge graph), combining semantic attention and label attention into GCN.
	
	\item MSGM\cite{lin2022semantic}: This method learns visual features using multi-grained semantic grouping mechanisms and directly extracts label correlations using the GCN module.
	
	\item ML-KNN\cite{zhang2007ml}: This method uses VGG16 for feature extraction and ML-KNN  for classification.
	
	\item MLRSSC-CNN-GNN\cite{lin2022semantic}: This method constructs a scene graph for each scene, where nodes of the graph are represented by super pixel regions of the scene. The multi-layer-integration graph attention network (GAT) model is proposed to fully mine the spatio-topological relationships of the scene graph.
	
	\item Gardner\cite{gardner2017multi}: This method uses a number of data augmentation and ensemble techniques.
	
	\item RBFNN\cite{zeggada2017deep}: This method uses the VGG16 for feature extraction and the RBFNN for classification.
	
	\item Stivaktakis\cite{stivaktakis2019deep}: This method is a data augmentation technique that can drastically increase the size of a smaller data set to copious amounts.
	
	\item CNN-RNN\cite{wang2016cnn}: This method learns a joint image-label embedding to characterize the semantic label dependency as well as the image-label relevance to integrate both information in a unified framework.
	
	\item Hunag\cite{huang2021multilabel}: This method fuse and refines the multiscale features from different layers of a CNN model by using a channel-spatial attention mechanism. Then, the label correlation information is extracted from a label co-occurrence matrix and embedded into the multiscale attentive features.
	
	\item Zhu\cite{zhu2020deep}: This method uses duallevel semantic concepts, where scene labels are used to guide multilabel classification.
	
\end{itemize}

As shown in Table \ref{tab:compare others}, our proposed method obtains a substantial improvement over baseline and surpasses the current advanced multi-label classification methods in remote sensing.
On the boneback of ResNet-50, the F1 score of the UCM data set improves by 12.56\% and reaches 93.96\%, the F1 score of the AID data set improves by 5.52\% and reaches 92.96\%, and the F1 score of the DFC15 data set improves by 4.33\% and reaches 97.46\%. The UCM data set and the AID data set use the same set of object. The UCM data set has relatively cleaner label relationships and fewer edges in the label graph, which makes it easier for SIGNA to infer the relationship between images and labels. This can be seen from Fig. \ref{fig:label matrix2}. After the transform of formula \ref{eq:smoothing}, the label co-occurrence graph of AID data set has a total of 92 unidirectional edges, while the label co-occurrence graph of UCM data set has only 38 unidirectional edges. This indicates that the label relationship complexity of the UCM data set is much lower than that of the AID data set.
Compared with ML-GCN, which also uses GCN for label-relational feature extraction, our method achieves 3.60\% and 3.38\% higher F1 scores on the UCM and AID data sets, respectively. This illustrates the better fusion of semantic relationship features and image features in our proposed SIGNA.

\subsection{Effectiveness Analysis}
\label{subsec:Effectiveness Analysis}

In Section \ref{subsub:Per-Class Case Studies}, we analyze the influence of the complexity of the label graph on the effectiveness of SIGNA by comparing the relationship between the improvement of the F1 score of the baseline for each class of SIGNA and the label graph. In Section \ref{subsub:Visualizing feature embedding using t-SNE}, we observe the impact of SIGNA networks on global features by Visualizing feature embedding using t-SNE. In Section \ref{subsub:Typical Case Analysis in Context Information Extraction}, we generate heatmaps for some typical examples through layerCAM\cite{jiang2021layercam}, analyze where the attention is focused after 
using SIGNA, and the impact of semantic relationships on classification.

\subsubsection{Per-Class Case Studies}
\label{subsub:Per-Class Case Studies}

As shown in Tables \ref{tab:UCM-per}, \ref{tab:AID-per}, and \ref{tab:DFC15-per}, the performance of the SIGNA and baseline on the UCM, AID and DFC15 data sets for each category is shown by F1, P, and R scores, respectively. In the following, we will analyze the differences in performance of each category on each data set in detail in conjunction with the label co-occurrence graphs (Fig. \ref{fig:label matrix2}).

Table \ref{tab:UCM-per} illustrates the performance of various classes of the UCM data set. Tanks, courts, cars, soils, and buildings have the largest improvement in F1 scores compared to baseline, with 16.67\%, 14.63\%, 9.77\%, 9.31\%, and 9.07\% improvement, respectively. Correspondingly, there are 3 edges, 5 edges, 9 edges, 10 edges and 10 edges on their label co-occurrence graphs (after formula \ref{eq:smoothing}), respectively.
In the UCM data set, cars, soil and buildings improved by 9.77\%, 9.31\% and 9.07\% respectively, while they only improved by only 0.22\%, 2.60\% and 1.59\%. in the AID data set. Corresponding to these three labels, the number of edges in the AID label co-occurrence graph is 17 for cars, 19 for soil, and 20 for buildings. While court (23.09\%), chaparral (20.03\%) and tanks (17.56\%), which have the largest F1 score improvement in the AID data set, have 6, 5 and 6 edges on the AID label co-occurrence graphs, respectively.

The relationship between the F1 score improvement and the number of label graph edges for the above two data sets also reaffirms our previous view. That is, the smaller the number of edges in the label co-occurrence graph, the more obvious the relationship between labels, and the larger the F1 score improvement using the SIGNA. 
This property is not only for the whole data set, but also for each category of the data set. And it is also clear from the comparison between the two data sets that the F1 score improvement is indeed brought about by the semantic relationship feature.

\begin{table}[htbp]
  \centering
  \caption{F1, P and R scores of our proposed method and baseline on each category of the DFC15 data set.}
    \begin{tabular}{|c|c|c|c|c|c|c|}
    \hline
    \hline
          & \multicolumn{3}{c|}{ours} & \multicolumn{3}{c|}{baseline} \bigstrut\\
    \hline
          & F1    & P     & R     & F1    & P     & R \bigstrut\\
    \hline
    impervious & \textbf{99.21 } & \textbf{99.40 } & \textbf{99.01 } & 98.61  & 98.81  & 98.42  \bigstrut[t]\\
    water & \textbf{97.20 } & \textbf{97.89 } & \textbf{96.53 } & 90.17  & 88.08  & 92.36  \\
    clutter & \textbf{95.91 } & \textbf{97.67 } & \textbf{94.21 } & 91.33  & 93.00  & 89.71  \\
    vegetation & \textbf{95.58 } & \textbf{98.78 } & \textbf{92.57 } & 91.22  & 90.45  & 92.00  \\
    building & \textbf{94.81 } & \textbf{96.69 } & \textbf{92.99 } & 87.50  & 85.89  & 89.17  \\
    tree  & \textbf{88.64 } & \textbf{90.70 } & \textbf{86.67 } & 79.01  & 88.89  & 71.11  \\
    boat  & \textbf{98.82 } & \textbf{100.00 } & \textbf{97.67 } & 78.57  & 80.49  & 76.74  \\
    car   & \textbf{95.28 } & \textbf{95.28 } & \textbf{95.28 } & 76.42  & 76.42  & 76.42  \bigstrut[b]\\
    \hline
    \hline
    \end{tabular}%
  \label{tab:DFC15-per}%
\end{table}%


\begin{table}[htbp]
  \centering
  \caption{F1, P and R scores of our proposed method and baseline on each category of the UCM data set.}
    \resizebox{.48\textwidth}{!}{
    \begin{tabular}{|c|c|c|c|c|c|c|}
    \hline
    \hline
          & \multicolumn{3}{c|}{ours} & \multicolumn{3}{c|}{baseline} \bigstrut\\
    \hline
          & F1    & P     & R     & F1    & P     & R \bigstrut\\
    \hline
    airplane & \textbf{100.00 } & \textbf{100.00 } & \textbf{100.00 } & 97.44  & 100.00  & 95.00  \bigstrut[t]\\
    soil  & \textbf{82.78 } & \textbf{81.70 } & \textbf{83.89 } & 73.47  & 74.48  & 72.48  \\
    buildings & \textbf{94.03 } & \textbf{92.65 } & \textbf{95.45 } & 84.97  & 74.71  & 98.48  \\
    cars  & \textbf{90.64 } & \textbf{92.81 } & \textbf{88.57 } & 80.88  & 89.58  & 73.71  \\
    chaparral & \textbf{97.67 } & \textbf{100.00 } & \textbf{95.45 } & 95.45  & 95.45  & 95.45  \\
    court & \textbf{97.56 } & \textbf{100.00 } & \textbf{95.24 } & 82.93  & 85.00  & 80.95  \\
    dock  & \textbf{100.00 } & \textbf{100.00 } & \textbf{100.00 } & 97.44  & 100.00  & 95.00  \\
    field & \textbf{100.00 } & \textbf{100.00 } & \textbf{100.00 } & 100.00  & 100.00  & 100.00  \\
    grass & \textbf{90.27 } & \textbf{88.83 } & \textbf{91.76 } & 84.73  & 76.79  & 94.51  \\
    mobile-home & \textbf{100.00 } & \textbf{100.00 } & \textbf{100.00 } & 95.00  & 95.00  & 95.00  \\
    pavement & \textbf{92.69 } & \textbf{91.63 } & \textbf{93.77 } & 92.53  & 86.99  & 98.83  \\
    sand  & \textbf{92.31 } & \textbf{90.00 } & \textbf{94.74 } & 85.22  & 84.48  & 85.96  \\
    sea   & \textbf{100.00 } & \textbf{100.00 } & \textbf{100.00 } & 100.00  & 100.00  & 100.00  \\
    ship  & \textbf{97.56 } & \textbf{100.00 } & \textbf{95.24 } & 97.56  & 100.00  & 95.24  \\
    tanks & \textbf{100.00 } & \textbf{100.00 } & \textbf{100.00 } & 83.33  & 93.75  & 75.00  \\
    trees & \textbf{91.99 } & \textbf{92.23 } & \textbf{91.75 } & 85.45  & 78.45  & 93.81  \\
    water & \textbf{98.77 } & \textbf{97.56 } & \textbf{100.00 } & 97.50  & 97.50  & 97.50  \bigstrut[b]\\
    \hline
    \hline
    \end{tabular}%
    }

  \label{tab:UCM-per}%
\end{table}%

\begin{table}[htbp]
  \centering
  \caption{F1, P and R scores of our proposed method and baseline on each category of the AID data set.}
    \begin{tabular}{|c|c|c|c|c|c|c|}
    \hline
    \hline
          & \multicolumn{3}{c|}{ours} & \multicolumn{3}{c|}{baseline} \bigstrut\\
    \hline
          & F1    & P     & R     & F1    & P     & R \bigstrut\\
    \hline
    airplane & \textbf{94.74 } & \textbf{94.74 } & \textbf{94.74 } & 85.71  & 93.75  & 78.95  \bigstrut[t]\\
    soil  & \textbf{85.10 } & \textbf{85.10 } & \textbf{85.10 } & 82.50  & 76.34  & 89.74  \\
    buildings & \textbf{96.37 } & \textbf{96.71 } & \textbf{96.03 } & 94.77  & 94.23  & 95.33  \\
    cars  & \textbf{94.29 } & \textbf{93.49 } & \textbf{95.10 } & 94.07  & 94.78  & 93.38  \\
    chaparral & \textbf{51.28 } & \textbf{52.63 } & \textbf{50.00 } & 31.25  & 41.67  & 25.00  \\
    court & \textbf{76.42 } & \textbf{83.93 } & \textbf{70.15 } & 53.33  & 60.38  & 47.76  \\
    dock  & \textbf{80.41 } & \textbf{81.25 } & \textbf{79.59 } & 75.27  & 79.55  & 71.43  \\
    field & \textbf{84.06 } & \textbf{90.63 } & \textbf{78.38 } & 67.61  & 70.59  & 64.86  \\
    grass & \textbf{95.38 } & \textbf{93.47 } & \textbf{97.37 } & 93.87  & 89.15  & 99.12  \\
    mobile-home & \textbf{0.00 } & \textbf{0.00 } & \textbf{0.00 } & 0.00  & 0.00  & 0.00  \\
    pavement & \textbf{98.06 } & \textbf{98.48 } & \textbf{97.63 } & 96.62  & 98.01  & 95.27  \\
    sand  & \textbf{91.59 } & \textbf{90.74 } & \textbf{92.45 } & 91.26  & 94.00  & 88.68  \\
    sea   & \textbf{98.82 } & \textbf{100.00 } & \textbf{97.67 } & 83.12  & 94.12  & 74.42  \\
    ship  & \textbf{76.77 } & \textbf{76.00 } & \textbf{77.55 } & 62.22  & 68.29  & 57.14  \\
    tanks & \textbf{97.56 } & \textbf{100.00 } & \textbf{95.24 } & 80.00  & 68.97  & 95.24  \\
    trees & \textbf{96.40 } & \textbf{94.94 } & \textbf{97.91 } & 94.24  & 92.90  & 95.62  \\
    water & \textbf{77.38 } & \textbf{87.41 } & \textbf{69.41 } & 67.57  & 79.37  & 58.82  \bigstrut[b]\\
    \hline
    \hline
    \end{tabular}%
  \label{tab:AID-per}%
\end{table}%

\subsubsection{Visualizing feature embedding using t-SNE}
\label{subsub:Visualizing feature embedding using t-SNE}

\begin{figure*}
    \centering

    \includegraphics[width=0.9\textwidth]{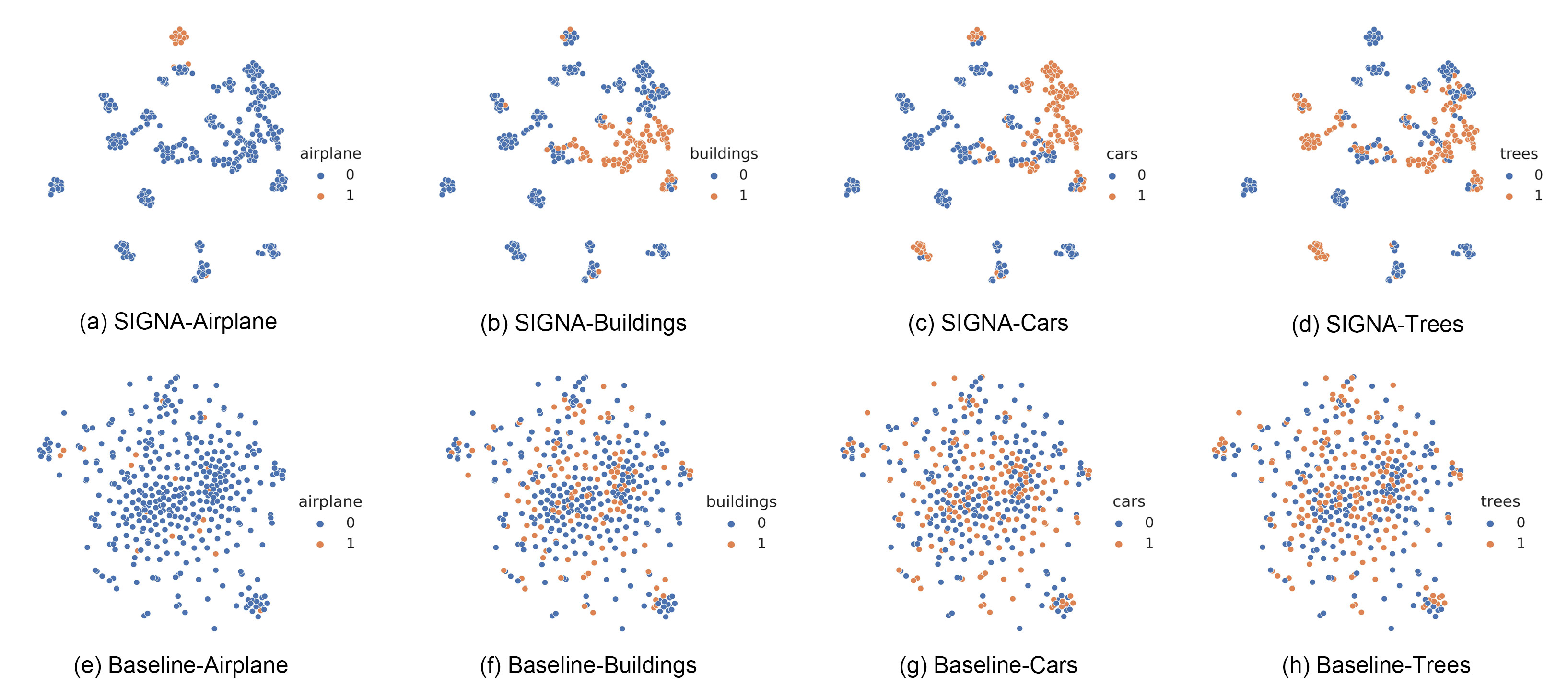} 
    \caption{Visualizing feature of SIGNA and baseline method using t-SNE on the UCM data set.}

    \label{fig:t-sne}
\end{figure*}

T-SNE\cite{van2008visualizing} is a feature dimensionality reduction method, which reduces high-dimensional features to two-dimensional, so that the feature structure can be visually represented on the figure.
As shown in Fig. \ref{fig:t-sne}, on the UCM data set, we use t-SNE to visualize the last layer feature maps of the baseline method and SIGNA. Fig. \ref{fig:t-sne} shows the feature distributions of the four categories of airplane, buildings, cars, and trees. Subfigures (a), (b), (c) and (d) show the feature distributions of the SIGNA, and subfigures  (e), (f), (g) and (h) show the feature distributions of baseline method.

Globally, the feature distribution figure shows that the features of SIGNA are clearly separated, while the features of baseline method are not. The difference becomes more obvious when the feature points of the four categories are labeled with different colors. For SIANA, the features of the same category are highly aggregated, and the features of different categories are clearly separated. For the baseline method, the features of the same category are highly dispersed, and different kinds of features are mixed together. 
This indicates that after SIGNA, the channels of the feature map are weighted by semantic relations, and the image features are guided to a feature space that is highly correlated with the semantics of the labels. 
The feature space of the baseline method is too far from the label semantics, so the feature distribution figure after t-SNE dimensionality reduction is confusing.
Observing subfigures (a) and (c), in (a) airplane category feature points are in the top pile, and in (c) cars category also have most of the feature points in the top pile. 
Observing subfigures (c) and (d), the trees category feature points in (d) are in the bottom left pile, while the cars category in (c) also have most of them in the bottom left pile. 
This illustrates that the feature aggregation is still good in the metric space constructed by the SIGNA, even though each image has multiple labels.

\subsubsection{Typical Case Analysis in Context Information Extraction}
\label{subsub:Typical Case Analysis in Context Information Extraction}
\begin{figure*}
    \centering
    \includegraphics[width=0.9\textwidth]{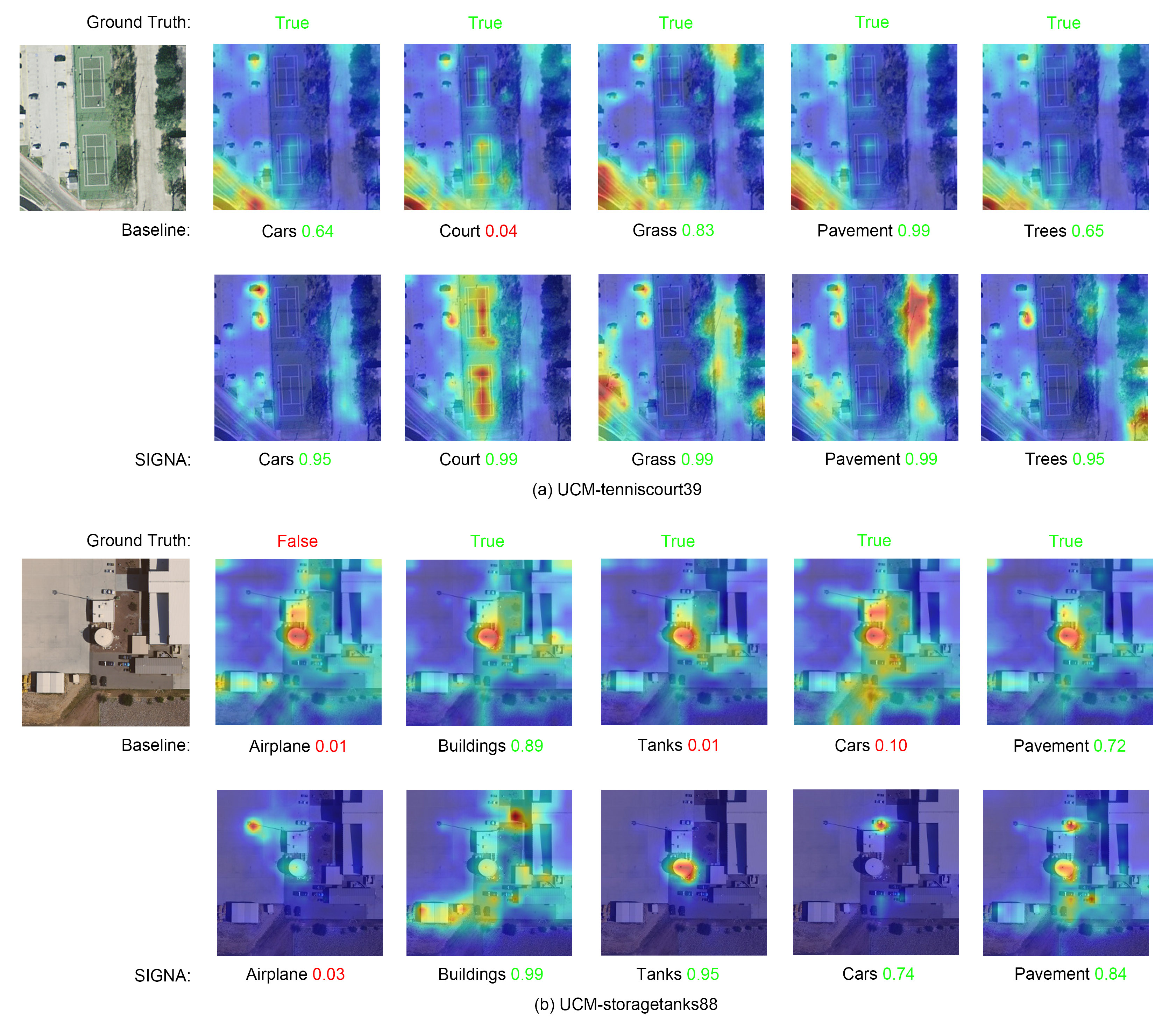} 
    \caption{The heatmap generated by the third layer feature map using layerCAM method for baseline mehtod and SIGNA}
    \label{fig:heatmap}
\end{figure*}

LayerCAM is used to generate heatmaps, whereby the attention regions of SIGNA can be observed.
LayerCAM collects object localization information from coarse to fine levels and integrates them into a high-quality class activation map where the pixels can associated with the object is better highlighted. 
With layerCAM, more can be observed when acting on shallow feature maps, and fine-grained details of the target object can be effectively preserved.
Confidence greater than 0.5 in Fig. \ref{fig:heatmap} means true, the label exists in the image. Confidence less than 0.5 means false, and the label does not exist in the image. 
As shown, we selected two images from the UCM data set and used layerCAM to map the first layer of features in the third layer of SIGNA and baseline method. The reason for choosing this layer for heat map plotting is that the SIGNA is inserted into the last layer of the baseline method second layer of the feature map, so that this layer can effectively observe the effectiveness of the SIGNA, what regions are of more interest, and whether label correlation can be exploited.

By observation, we can learn that, first, in (a) and (b), for each classification, the attention distribution of ResNet-50 is similar, but the attention distribution of SIGNA differs more. 
Second, after passing SIGNA, the network is able to pay more attention to small object features, such as cars and trees in (a) and cars and tanks in (b). 
Compared to the baseline method, the network focuses the otherwise distracting attention on a very small part, such as the car and tank object parts.
This also illustrates the ability of SIGNA to stimulate or inhibit the correct or incorrect channel to keep their attention tightly locked on the correct object.

Looking at subgraph (a), in the baseline method, the classification of the court is wrong, while in SIGNA, the classification of the court is correct.
In the label co-occurrence map of the UCM data set, the points connected to the court are buildings, cars, grass, pavement and trees. And the baseline method correctly classified cars, grass, pavement and trees. 
We can speculate that through SIGNA, the network knows that all four points associated with the court exist, inferring that the court may also exist with high probability. Thus the channels associated with the court are weighted.
Looking at subgraph (b), in the baseline method, the classification of tanks and cars are wrong, while in SIGNA, the classification of tanks and cars are correct.
In the label graph of UCM data set, the points connected with tanks are buildings, soils, and pavements. 
And the classification of cars and pavements by baseline method is correct. 
We can speculate that through SIGNA, the network knows that two points associated with tanks exist, inferring that tanks may also exist with high probability. Thus the channels associated with the tanks are weighted.

\section{Conclusion and Future Work}
\label{sec:Conclusion and Future Work}

\subsection{Conclusion}
\label{subsec:Conclusion}
In this paper, we propose a label correlation-based channel attention mechanism for MLRSIC: Label Graph Channel Attention Networks. 
The method utilizes GNN to generate relational features between labels, and then uses semantic interleaving coding to guide image features into the feature space related to semantic relations. Global channel attention is implemented in this feature space.
Experimental results on three multi-label data sets, UCM, AID and DFC15, show that the proposed SIGNA exhibits the best performance compared to popular multi-label image classification methods.

\subsection{Future Work}
\label{subsec:Future Work}

\begin{itemize}
\item Explore more ways to encode label relationships. In this paper, the label co-occurrence matrix is used to explicitly construct the graph, and the Glove word embedding of the label is used as the input of the GNN. The relationship between the word embedding of the label and the label is weakly correlated, we can try to find a better GNN input, and the correlation with the label relationship is stronger.
\item Explore multi-level feature fusion methods. The SIGNA in this paper can be inserted into any layer of CNN to function, and the fusion effect of label relationship feature and image feature is better. From a global perspective, from low-level to high-level, feature fusion can be performed to achieve the effect of using low-level texture features and high-level abstract features at the same time.
\end{itemize}

\ifCLASSOPTIONcaptionsoff
  \newpage
\fi

\bibliographystyle{IEEEtran}
\bibliography{bibfile}

\end{document}